\documentclass{article}

\usepackage{amsmath}
\usepackage[utf8]{inputenc} 
\usepackage[T1]{fontenc}    
\usepackage{hyperref}       
\usepackage{url}            
\usepackage{booktabs}       
\usepackage{amsfonts}       
\usepackage{nicefrac}       
\usepackage{microtype}      
\usepackage{lipsum}
\usepackage{xcolor}
\usepackage{color,soul}
\usepackage{amsmath}
\usepackage{amsthm}
\usepackage{amssymb}
\usepackage{amsfonts}
\usepackage{bbm}
\usepackage{bm}
\usepackage{mathtools}
\usepackage{graphicx}
\usepackage{epsfig,epsf,url,enumitem}
\usepackage{accents}
\usepackage{svg}
\usepackage{longtable}
\usepackage{amsbsy}
\usepackage[noadjust]{cite}
\usepackage{authblk}

\usepackage{thmtools,thm-restate}

\newtheorem{theorem}{Theorem}

\theoremstyle{definition}

\usepackage{algorithm}
\usepackage{algorithmic}

\newcommand{\Expxt}[1]{\mathbb{E}_{x_0 \sim \mathcal{D}, \omega(0)\sim \rho} \left[ #1 \right]}

\newcommand{\1}[1]{\mathbf{1}_{#1}}
\newcommand{\eps}{\varepsilon}

\newcommand{\sumtinf}{\sum_{t=0}^\infty}

\newcommand{\cD}{\mathcal{D}}

\newcommand{\cT}{\mathcal{T}}

\newcommand{\cF}{\mathcal{F}}

\renewcommand{\it}{\ell}

\newcommand{\E}{\mathbb{E}}

\usepackage{newfloat}
\usepackage{listings}
\usepackage{graphicx}

\title{Global Convergence Using Policy Gradient Methods for Model-free Markovian Jump Linear Quadratic Control}
\author {
    \textsuperscript{\rm *, 1} Manoj Bhadu,
    \textsuperscript{\rm *, 1} Santanu Rathod, 
    \textsuperscript{\rm 1} Abir De
}

\date{%
    $^1$IIT-Bombay \\%
    $^*$equal contribution \\[2ex]%
    \today
}
\begin{document}
\maketitle

\begin{abstract}
Owing to the growth of interest in Reinforcement Learning in the last few years, gradient based policy control methods have been gaining popularity for Control problems as well. And rightly so, since gradient policy methods have the advantage of optimizing a metric of interest in an end-to-end manner, along with being relatively easy to implement without complete knowledge of the underlying system. In this paper, we study the global convergence of gradient-based policy optimization methods for quadratic control of discrete-time and \textbf{model-free} Markovian jump linear systems (MJLS). We surmount myriad challenges that arise because of more than one states coupled with lack of knowledge of the system dynamics and show global convergence of the policy using gradient descent and natural policy gradient methods. We also provide simulation studies to corroborate our claims. 
\end{abstract}
\section{Introduction}
Oftentimes, in reality, control systems don’t behave as they’re theoretically modelled. And sometimes the changes can be too abrupt for the system to be expected to observe a fixed prior behaviour, component failures and repairs or for example environmental disturbances or
changes in subsystems interconnections or a change in network based models, like air-transportation or disease epidemics \cite{Salathe et al.}, due to some confounders, etc. In some cases these systems can be modelled by discrete-time linear systems with state-transition coming from an underlying Markov chain; like for instance in the case of ship steering \cite{Astrom et al.} the ship dynamics vary according to the speed, which can be measured from appropriate speed sensors, and autopilots for these systems can be improved by taking these changes into account.
MJLS as a class of control problem has been widely studied, \cite{Fox et al.}, \cite{Sworder et al.}, \cite{Shalom et al.}, with wide array of practical applications like \cite{Gopalakrishnan et al.} modelling the time-varying networks and switching topologies of networked systems such as air-transportation via discrete-time, positive Markov Jump Linear Systems. Often in these systems the jump probabilities are assumed to be known, although not when, if at all, the jumps will occur \cite{Fragoso et al.}, in subsequent sections we entertain cases without this assumption as well. Another main question that arises is of whether or not the  operation mode $\omega(t)$ is known, and oftentimes it can be known by placing appropriate sensors. Having said that \cite{Vargas et al.} explores the scenario of no mode observation in MJLS. We in this work assume that the operation mode $\omega(t)$ can be observed. \\\\
Also, knowing the system dynamics before hand might not be the fastest and autonomous way to go about finding optimal policy, especially in systems where the states alter dynamically. And in this paper we've shown that optimal policy can simply be achieved by gathering data and statistics, and hence it's some sense more autonomous. \cite{Kakade et al.} show model-free convergence guarantees for LQR systems, but the results can't be trivially extended to MJLS because of the dynamic switch between states instead of being static. \cite{Porto et al.} explores MJLS policy convergence when the system parameters are known, and thus our paper tries to bridge this gap between model-given MJLS \cite{Porto et al.} and model-free LQR \cite{Kakade et al.}

\subsection{Our work}
We start by mentioning notations and conditions that we use throughout the paper, and then write down Algorithm-1 which gives us a procedure to estimate gradients and state correlation matrix without knowing anything about the state dynamics. Before stating the main policy convergence convergence results we convey some intermediary key ideas that help us get there. 
\\
Primarily, we establish that a. For sufficient time horizon the cost function and state-correlation matrix can estimated near accurately, b. That slight perturbations in policy doesn't results in large variations in estimated gradients or correlation matrix and the variations are bounded. We also show that for a given policy the estimates of gradients and correlation matrix are bounded when then transition probability matrix differ by a small $\epsilon_{p}$.These intermediary results along with a few other arguments mentioned in the Appendix section then help up prove policy convergence using gradient descent and natural gradient descent which is our main result. 
\subsection{Summary of contributions}
\begin{itemize}
    \item We provide an algorithm for estimating gradient and state correlation matrix used in gradient descent and natural gradient descent for policy iteration in model-free MJLS case.
    \item We prove intermediary results showing that a sufficient time horizon is enough for a good estimated of concerned quantities, and that variations resulting due slight perturbations in policy and transition probabilities can be bounded.
    \item We then prove policy convergence using both gradient descent and natural gradient techniques.
    \item We also provide simulation studies to corroborate our convergence claims.
\end{itemize}

\subsection{Related works}
Recently there have been several advances in deep Reinforcement Learning, like AlphaGo \cite{Silver et al.} with an over-arching goal of a general learning agent, but those techniques are not theoretically well understood and as such this lack of firm theoretical guarantees might prove disastrous as ML systems get more ambitious. And hence there’s been growth in studies trying to solve well studied control problems like LQR using policy gradient techniques \cite{Kakade et al.} which are comparatively easier to analyse and provide 
provable convergence guarantees of the optimal policy. And as an extension one of our primary motivations to study MJLS from a model-free policy gradient perspective 
was understand the theoretical challenges better more generally, since LTI LQR is but a special case of MJLS LQR with $N_{s}=1$ or just a single operating mode. \cite{Porto et al.} studies the MJLS problem from a model-based point of view, and provide novel Lyapunov arguments regarding the stability of gradient based policy iterations and its convergence there-of. As ML evolves to tackle more ambitious tasks it’s imperative that systems are able to progress and learn using data-driven approaches on the go, which calls for us to pay attention to model-free settings where one doesn’t have access to system/cost dynamics and is yet supposed to obtain the optimal policy using whatever sampled data is available from the environment. And along this line of thinking, apart from \cite{Kakade et al.}, there have been studies like \cite{Dean et al.} which solve the optimal LQR contoller when the dynamics are unknown using a multi-stage procedure, procedure called Coarse-ID control, that estimates a model from a few experimental trials, estimates the error in that model with respect to the truth. \cite{Bradtke et al.} follows a DP approach, wherein the  Q-function is estimated using RLS(recursive least squares) and is used to prove convergence. Unlike LQR model-based control settings which have been well studied classically \cite{Ljung et al.}, \cite{Claude et al.} or with \cite{Yasin et al.} using bandit approaches or \cite{Hazan et al.} studying it in an online manner, extension of these techniques to model-free settings hasn’t been as prolific due to inherent difficulties one faces when one doesn’t know about system dynamics. And studying MJLS LQR without the knowledge of system dynamics is harder still due to the jump-parameter $\omega(t)$, and the coupling between state/input matrices and $\omega(t)$. And thus finding proper gradient estimates and proving convergence in MJLS setting becomes a non-trivial task. To that end we provide results in following sections to tackle those problems.

\section{Preliminaries and Background}
\subsection{Notation}
We denote the set of real numbers by $\mathbb{R}$. Let $A$ be a matrix, then we use the notation $A^{\top}$, $\|A\|$, tr($A$), $\Gamma_{\min}(A)$, and $\rho(A)$ to denote its transpose, maximal singular value, trace, minimum singular value, and spectral radius respectively. Given matrices $\{D_i\}_{i=1}^{m}$, let diag($D_1,\cdots, D_m$) denote the block diagonal matrix whose ($i,i$)-th block is $D_i$. We use vec($A$) to denote the vectorization of matrix $A$. The positive definiteness and positive semi-definiteness of symmetric matrix $Z$ is denoted by $Z \succ 0$ and $Z \succeq 0$. \\\\
We now define some MJLS literature motivated notations along with operators/notations that we'll be using in this paper. Let $\mathbb{M}^{N_{s}}_{n\times m}$ denote the space made up of all $N_s$- tuples of real matrices $V$= ($V_1,\cdots,V_{N_{s}}$) with $V_{i} \in \mathbb{R}^{n \times n}$, $i \in \mathbb{N}$. For $V$= ($V_1,\cdots,V_{N}$) $\in \mathbb{M}^{N}$ we define:

\begin{align*}
&\|V\|_{1}:= \sum_{i= 1}^{N_{s}} \|V_{i}\|, \hspace{10pt} \|V\|_{2}^{2}:= \sum_{i=1}^{N}tr(V_{i}^{\top}V_{i}) \\
\|V\|_{\max}:=&\max_{i=1,\cdots,N}\|V_{i}\|, 
\Lambda_{\min}(V):= \min_{i=1,\cdots,N}\sigma_{\min}(V_{i}),\\
&trace(V) = \sum_{i=1}^{N_s} tr(V_i)
\end{align*}

For $V, S \in \mathbb{M}^{N}$, their inner product is defined as: \\
\[
\langle V, S \rangle := \sum_{i=1}^{N} tr(V_{i}^{\top}S_{i})
\]
We now introduce some new operators: $\{ \cF_K(v), \cT_{K}(V)\}$ that make analysis easy, will be used while proving intermediary results. 
\[
\cF_K(V) = (\cF_{K,1}(V), \ldots, \cF_{K,Ns}(V))
\]
\[
\cF_{K,j}(V) = \sum_{i=1}^{n_s} p_{ij} (A_i-B_i K_i)V_i(A_i-B_i K_i)^\top \, 
\]
\[
\cT_{K}(V) =\sum_{t=0}^{\infty} \cF^{t}_{K}(V), 
\]


\subsection{Markovian Jump Linear Systems}
A Markovian jump linear system (MJLS) is governed by the following discrete-time state-space model

\begin{equation} \label{eq:1}
    x_{t+1}= A_{\omega(t)}x_{t}+B_{\omega(t)}u_{t}
\end{equation}

where $x_{t} \in \mathbb{R}^{d}$ is the system state, and $u_{t} \in \mathbb{R}^{k}$ corresponds to the control action. The initial state $x_{0}$ is assumed to have a distribution $D$. The system matrices $A_{\omega(t)} \in \mathbb{R}^{d \times d}$ and $B_{\omega(t)} \in \mathbb{R}^{d \times k}$ depend on the switching parameter $\omega(t)$, which takes values on $\Omega:= \{1, \cdots, N_{s}\}$. We will denote $A$= ($A_1, \cdots, A_{N_{s}}$)$\in \mathbb{M}^{N_{s}}_{d \times d}$ and $B$= ($B_{1}, \cdots, B_{N_{s}}$)$\in \mathbb{M}^{N_{s}}_{d \times k}$.
The jump parameter $\{\omega(t)\}_{t=0}^{\infty}$ comes from a time homogenous Markov chain whose transition probability is given as 

\begin{equation} \label{eq:2}
    p_{ij}= \mathbb{P}(\omega(t+1)=j| \omega(t)=i)
\end{equation}
Let $P$ be the probability transition matrix whose ($i,j$)-th entry is $p_{ij}$, where $p_{ij} \geq 0$ and $\sum_{j=1}^{N_{s}}p_{ij}= 1$. The initial distribution of $\omega(0)$ is given by $\pi = [\pi_{1}, \cdots, \pi_{N_{s}}]^{\top}$, and we have $\sum_{i= 1}^{N_{s}} \pi_{i}=1$. Moreover we assume that \ref{eq:1} can be mean square stabilized. \\
For our work we focus on quadratic optimal control where the objective is to choose control actions $\{u_{t}\}_{t=0}^{\infty}$ to minimize the following cost function:
\begin{equation} \label{eq:3}
    C= \mathbb{E}_{x_{0} \sim D, \omega_{0} \sim \pi} \left[ \sum_{t=0}^{\infty} x_{t}^{\top}Q_{\omega(t)}x_{t}+u_{t}^{\top}R_{\omega(t)}u_{t} \right]
\end{equation}
Throughout the paper it's assumed that $Q= (Q_1, \cdots, Q_{N_{s}}) \succ 0$, and $R= (R_1, \cdots, R_{N_{s}}) \succ 0$, $\pi_{i}> 0$ so that there's non-zero probability of the system starting from a particular state, and $\mathbb{E}_{x_{0}\sim D}[x_{0}x_{0}^{\top}] \succ 0$ so that the  expected covariance of the initial state is full-rank. The assumptions are quite standard in the learning-based control and can be surmised as the persistently excitation condition in the system identification literature. The above problem can be called as "MJLS LQR" as mentioned in \cite{Porto et al.}. The optimal controller for this problem is defined by dynamics \ref{eq:1}, transition probabilities \ref{eq:2}, and cost \ref{eq:3} can be obtained by solving a system of coupled Algebraic Riccati Equations (AREs) \cite{Fragoso et al.} \\

Now, it is well known that the optimal cost can be achieved by a linear feedback of the form, 
\begin{equation}\label{eq:4}
    u_{t}= -K_{\omega_{t}}x_{t}
\end{equation}
with $K= (K_1, \cdots, K_{N_{s}}) \in \mathbb{M}^{N_{s}}_{k \times d}$. Combining the linear policy \ref{eq:4} with \ref{eq:1}, we obtain the closed-loop dynamics: 
\begin{equation} \label{eq:5}
    x_{t+1}= (A_{\omega(t)}-B_{\omega(t)}K_{\omega(t)})x_{t}= \Gamma_{\omega(t)}x_{t}
\end{equation}
with $\Gamma= (\Gamma_{1}, \cdots,\Gamma_{N_{s}}) \in \mathbb{M}^{N_{s}}_{d \times d}$. Note that using \ref{eq:4} we can rewrite the cost \ref{eq:3} as, 
\begin{align*}
    C= \mathbb{E}_{x_{0} \sim D, \omega_{0} \sim \pi} \left[ \sum_{t=0}^{\infty}x_{t}^{\top}(Q_{\omega(t)}+K_{\omega(t)}^{\top}R_{\omega(t)}K_{\omega(t)})x_{t} \right]
\end{align*}

And for finding the optimal policy $\{K^{*}_{i}\}_{i \in \Omega}$ for $i \in \Omega$, we first define an operator $\mathcal{E}: \mathbb{M}^{N_{s}}_{d \times d} \rightarrow \mathbb{M}^{N_{s}}_{d \times d}$ as $\mathcal{E}(V):=(\mathcal{E}_{1}(A), \cdots, \mathcal{E}_{N_{s}}(V))$ where $V= (V_{1}, \cdots, V_{N_{s}}) \in \mathbb{M}^{N_{s}}_{d \times d}$ and $\mathcal{E}_{i}(V):= \sum_{j=1}^{N_{s}}p_{ij}V_{j}$. Now let $\{P_{i}\}_{i \in \Omega}$ be the unique positive definite solution to the following AREs: 

\begin{equation} \label{eq:6}
\begin{aligned}
    P_i = Q_i &+  A_i^T \mathcal{E}(P) A_i - A_i^T \mathcal{E}(P) B_i \times \\
&\left( R_i + B_i^T \mathcal{E}(P) B_i \right)^{-1} B_i^T \mathcal{E}(P) A_i.
\end{aligned}
\end{equation}
It can then be shown that the linear state feedback controller minimizing the cost function \ref{eq:3} is given by

\begin{equation} \label{eq:7}
    K^{*}_{i}= (R_{i}+B_{i}^{\top}\mathcal{E}_{i}(P)B_{i})^{-1}B_{i}^{\top}\mathcal{E}_{i}(P)A_{i}
\end{equation}

\subsection{Problem Setup}
In this section we reintroduce the policy optimization reformulation of the MJLS problem as was done in \cite{Porto et al.} and elaborate on its differences with our model-free setting. We also introduce a few additional operators which are useful while proving certain results we use in later sections. 

\paragraph{Policy Optimization problem for model-based MJLS:}

\begin{align*}
    \textit{minimize :} & \textit{ cost($C(K)$), given in \ref{eq:3}}\\
    \textit{subject to :}  & \textit{ a. state dynamics, given in \ref{eq:1}}\\
    &\vspace{10pt} \textit{b. control actions given in \ref{eq:4}}\\
    &\vspace{20pt} \textit{c. transition probabilities, given in \ref{eq:2}}\\
    &\vspace{10pt} \textit{d. K stabilizes \ref{eq:3} in the mean square sense}
\end{align*}

Let $\mathcal{K}$ denote the set of feasible set of stabilizing policies.
We know that for a given $K$, the resultant closed-loop MJLS \ref{eq:5} is mean square stable(MSS) if for any initial condition $x_{0} \in \mathbb{R}^{d}$ and $\omega(0) \in \Omega$ we get $\mathbb{E}[x_{t}x_{t}^{\top}] \rightarrow 0$ as $t \rightarrow \infty$ \cite{Costa et al.}, and thus we can trivially expand MSS arrive at the conclusion that $C(K)$ is finite if and only if $K$ stabilizes closed-loop dynamics in the mean square sense or if $K \in \mathcal{K}$. \\
It is useful to see how cost gradients are calculated in model-based settings to appreciate why we can't use similar methods and resort to estimating gradients using zeroth order optimization methods \cite {Flaxman et al.} shown later. For $K \in \mathcal{K}$, the cost $C(K)$ is finite and differentiable, and we can rewrite the cost \ref{eq:3}
as:
\begin{equation}\label{eq:8}
    C(K)= \mathbb{E}_{x_{0} \sim \mathcal{D}} \left[ x_{0}^{\top}\left(\sum_{i= 1}^{N_{s}} \pi_{i}P^{K}_{i} \right)x_{0} \right]
\end{equation}
Where $\pi$ is the initial distribution and $\{P^{K}_{i}\}_{i \in \Omega}$ are the solution to the coupled Lyapunov equations

\begin{equation} \label{eq:9}
    P_{i}^{K}= Q_{i}+ K_{i}^{\top}R_{i}K_{i}+(A_{i}-B_{i}K_{i})^{\top}\mathcal{E}_{i}(P^{K})(A_{i}-B_{i}K_{i})
\end{equation}
    Where $\mathcal{E}_{i}(P^{K})= \sum_{j=1}^{N_{s}}p_{ij}P_{j}^{K}$ as defined earlier. We also a define a new variable $X(t)= (X_{1}(t), \cdots, X_{N_{s}}(t))$ -we'll use it throughout the paper- such that $X_{i}(t):=\mathbb{E}[x_{t}x_{t}^{\top}\mathbbm{1}_{\omega(t)=i}]$ and this matrix also satisfies the recursion relation:

\begin{equation} \label{eq:10}
    X_{j}(t+1)= \sum_{i= 1}^{N_{s}}p_{ij}(A_{i}-B_{i}K_{i})X_{i}(t)(A_{i}-B_{i}K_{i})^{\top}
\end{equation}
owing to \ref{eq:5}, with $X_{i}(0)= \pi_{i}\mathbb{E}_{x_{0} \sim \mathcal{D}}[x_{0}x_{0}^{\top}], \forall i \in \Omega ,$ being the value at $t=0$. Now from \textbf{Lemma 1} in \cite{Porto et al.} we know that for $K \in \mathcal{K}$ the gradient for cost-function \ref{eq:3} wrt policy $K$ is:
\begin{equation} \label{eq:11}
    \nabla C(K)= 2 [L_{1}(K), L_{2}(K) \cdots L_{N_{s}}]\chi_{K}
\end{equation}
where $L_{i}(K)= (R_{i}+B_{i}^{\top}\mathcal{E}_{i}(P^{K})B_{i})K_{i}-B_{i}^{\top}\mathcal{E}_{i}(P^{K})A_{i}$ and\\
$\chi_{K}=$ $diag\left( \sum_{t=0}^{\infty}X_{1}(t), \cdots, \sum_{t=0}^{\infty}X_{N_{s}}(t)\right)$ \\
One can see how the empirical estimates of gradient from \ref{eq:11} can't be used since it requires information about the model, and this is partly what also makes the model-free problem challenging. \\
\paragraph{Policy Optimization problem for model-free MJLS:}
For the model-free MJLS setting we only have a black-box access to the system, meaning that the state ($x_{t}$) still evolves according to \ref{eq:1} and the cost is still accrued according to \ref{eq:3} only we don't know the system variables $\{A_{i}\}, \{B_{i}\}$ and cost-variables $\{Q_{i}\}, \{R_{i}\}$. Or formally we have,
\begin{align*}
    \textit{minimize :} & \textit{ cost($C(K)$), given in \ref{eq:3}}\\
    \textit{subject to :}  & \textit{ a. $\{A_{i}\}_{i \in \Omega}, \{B_{i}\}_{i \in \Omega}$ from the state dynamics \ref{eq:1}} \\
    & \textit{are not known}\\
    & \textit{b. $\{Q_{i}\}_{i \in \Omega}, \{R_{i}\}_{i \in \Omega}$ from the cost-function \ref{eq:3}}\\
    & \textit{are not known}\\
    & \textit{c. control actions given in \ref{eq:4}}\\
    &\textit{d. K stabilizes \ref{eq:3} in the mean square sense}\\
    &\textit{e. For the transition probabilities, given in \ref{eq:2}, we have}\\
    &\textit{two cases, when $p_{ij}$ is given and when it's not.}
\end{align*}

As seen from the constraints above we have to devise empirical estimates of gradient and other relevant parameters without the knowledge of model and cost parameters. We'll elaborate more on that in the following sections. We also need to make sure that our policy iterations coming from empirical estimates don't wander into the region of unstability while converging towards optimal policy. We'll expand on \textbf{Lemma 4} in \cite{Porto et al.}, which shows that policy update $\hat{K}^{'}$ from a stabilizing policy $\hat{K}$ using natural gradient with appropriate step-size $\eta$ is also stabilizing, to show the stability of our policy updates based on empirical estimates.

\section{The Algorithm}
In this section we present our gradient estimation algorithms and discuss the intermediary results required to achieve convergence. 
For data driven (or model free) approaches in Reinforcement Learning or control in general a standard way is to somehow estimate the terms of interest using simulations, \cite{Bradtke et al.}, \cite{Bertsekas et al.} and then show that
if the simulations are run for a sufficient amount of time then the estimated term resembles the actual term to a large extent, using some sort of concentration or other
inequality results. We elaborate on the above points further in the following subsections. 
\subsection{Description of Algorithm-1}
\begin{algorithm}
	\caption{ Model-Free Policy Gradient (and Natural Policy
          Gradient) Estimation for MJLS}
	\label{model_free_MJLS}
	\begin{algorithmic}[1]
	    \IF {Probability transition matrix $P$ = $[p]_{ij}$ is known}
        \STATE  we have {n}=0, with $P_{\text{inp}} = P$
        \ELSE
        \STATE Let, 
        \begin{align*}
            n_{1} &=\dfrac{1}{\epsilon^{2}\pi_{*}}\max\{d, \ln{\dfrac{1}{\epsilon \delta_{p}}}\} \\
            n_{2} &= \dfrac{1}{\gamma_{ps} \pi_{*}}\ln(\dfrac{d \| \mu/\pi\|_{2, \pi}}{\delta_{p}})
        \end{align*}
        \STATE Then $n = c\max \{ n_{1}, n_{2} \}$
        \STATE Now let $ \omega(1), \cdots, \omega(n) \sim P$ be the be the state values sampled from the Markov chain, and $N_{i}=\sum_{t=1}^{n-1}\mathbb{I}\{\omega(t)=i\}$, $N_{ij}=\sum_{t=1}^{n-1}\mathbb{I}\{\omega(t)=i,\omega(t+1)=j\}$ and so we have,
        \STATE $P_{\text{inp}}= [\hat{p}]_{ij}$ with $\hat{p}_{ij}= \dfrac{N_{ij}}{N_{i}}$ when $N_{i} \neq 0$ and $1/d$ when $N_{i}=0$
        \ENDIF
        \STATE \textbf{Input}: $\hat{K_{0}}$, $P_{\text{inp}}$,$\widehat{\nabla C(\hat{K_{0}})} = [0]_{ij}$, number of trajectories $m$, roll out length $\ell$, smoothing parameter $r$, dimension $d$, step-size $\eta$, error-tolerance $error$
        \FOR {$t=n+1$ to $\infty$}
		\FOR{$i = 1, \cdots m$}
		 \STATE Choose $U_i$ s.t. it is
                drawn uniformly at random over matrices whose (Frobenius) norm is $r$
        \STATE choose a starting state  $x_0\sim D$
		\FOR{$j = 1, \cdots l$}
		\STATE Choose K from $\hat{K}_{t-n}$ with transition probability and initial probability $[p_{\text{inp}}]_{ij}$ and $\rho_j$ respectively
		\STATE Sample a policy $\Tilde{K}_j = K+U_i$
        \STATE Simulate $\tilde{K}_j$. Now we calculate empirical estimates as follows:
\[
\widehat C_{i,\omega(j)} \mathrel{+}=   \frac{U_i c_j}{r^2} \, , \quad \widehat X_{i,\omega(j)} \mathrel{+}= x_j x_j^\top
\]
where $c_j$ and $x_j$ are the costs and states on this trajectory.
        \ENDFOR
		\ENDFOR
		\STATE Get the (biased) estimates:
		
\begin{align*}
&\widehat{\nabla C(K)} = \frac{1}{m} \sum_{i=1}^m d\ \widehat C_i \\
&\widehat \chi_K = \frac{1}{m} \sum_{i=1}^m\textup{diag}\left(X_{i,1},\cdots,X_{i,N_s}\right)
\end{align*}
\STATE We thus update the policy estimate as: 
\STATE $\widehat{K}_{t- n+1}= \widehat{K}_{t-n}- \widehat{\nabla C(K)}\widehat{\chi_{K}^{-1}}$
\IF {$\| \widehat{K}_{t- n+1}- \widehat{K}_{t-n}\| < error$}
\STATE Return $\widehat{K}_{t- n+1}$
\ENDIF
\ENDFOR
	\end{algorithmic}
\end{algorithm}

We know from \cite{Porto et al.} that for natural policy gradient method, the policy update rule is given by:
\begin{equation}\label{eq:12}
\hat{K}^{n+1}= \hat{K}^{n}- \eta \nabla C(\hat{K}^{n}) \chi^{-1}_{\hat{K}^{n}}
\end{equation}
And from \ref{eq:11} we know that cost-gradient $\nabla C(\hat{K}^{n})$ depends on state dynamics $\{ A_{i}, B_{i}, Q_{i}, R_{i} \}_{i \in N_{s}}$ and thus we can't exactly use empirical estimates of \ref{eq:11}. To that end we use zeroth-order optimization (Lemma \ref{lemma:stokes}) also present in \cite{Flaxman et al.} where in the cost-function-gradients can be estimated by using appropriate cost-functions values only. \\

From \textbf{Algorithm-1} we essentially want estimates of two quantities, a. gradient $\nabla C(K)$, b. state-correlation matrix $\chi_{K}$, and use those estimates during each policy iteration (Line 22 of \textbf{Algorithm-1})
\\\\
In the initial part of the algorithm depending on whether or not the transition probability matrix is given we either use the provided transition probability matrix or estimate it via sampling a Markov chain, the appropriate $P_{\text{inp}}$ is then used for rolling out simulated trajectories.
\\\\
As can be seen in \textbf{Algorithm-1} we input current policy estimate $\hat{K}$, number of trajectories $m$ which is nothing but the number of parallel trajectories we average on, roll out length $l$ which is the time-horizon for each of our $m$ trajectories, $r$ which bounds the Frobenius norm of our random matrices $\{U_{i}\}_{i \in [m]}$ used for zeroth-order optimization, and state dimension $d$. \\
For each trajectory $i$, as given in the \textbf{Algorithm-1}, based on the operating-mode $\omega(j)$ we accumulate cost $c_{j}$ for appropriate $\widehat{C}_{i, \omega(j)}$ for each time instance $j$ and also modify the correlation matrix $\widehat{X}$ accordingly. At the end of all the $m$ trajectories we finally get the empirical $\nabla \widehat{C(K)}$ and $\widehat{X}^{K}$ to be used for next policy iteration \ref{eq:12} with proper step-size $\eta$. And this process continues till convergence, we'll validate the choice our empirical estimates shortly.

\subsection{Estimating the transition probability matrix}

The estimator for the transition probability matrix of a Markov chain-- from its observed samples $ \omega(1), \cdots, \omega(n) \sim P$ and number of state $i$ samples  $N_{i}=\sum_{t=1}^{n-1}\mathbb{I}\{\omega(t)=i\}$ and number of $ij$ state transitions $N_{ij}=\sum_{t=1}^{n-1}\mathbb{I}\{\omega(t)=i,\omega(t+1)=j\}$-- can be written as: 
\begin{equation}\label{eq:13}
    P_{\text{inp}}= [\hat{p}]_{ij}
\end{equation}
with $\hat{p}_{ij}= \dfrac{N_{ij}}{N_{i}}$ when $N_{i} \neq 0$ and $1/d$ when $N_{i}=0$. \\

While the estimator is fairly simple to write down proving tight bounds and sample complexity guarantees are not very straightforward because of the presence of random variables in both the numerator and denominator or the transition probability estimate $\hat{p}_{ij}= \dfrac{\sum_{t=1}^{n-1}\mathbb{I}\{\omega(t)=i, \omega(t+1)=j\}}{\sum_{t=1}^{n-1}\mathbb{I}\{\omega(t)=i\}}$
\\\\

\cite{Kontorovich et al.} circumvent this complexity in their Theorem 3.1 by first fixing the denominator and using Marton Coupling arguments to bound the numerator, and then bound the probability of the denominator lying in some region. We write the Theorem 3.1 from \cite{Kontorovich et al.} below.

\begin{theorem}\label{theorem: markov_chain_est}
(Sample complexity upper bound w.r.t $\|.\|_{\infty}$ when $|\Omega| < \infty$) Let $\epsilon \in (0,2), \delta_{p} \in (0,1),$ and let 
$\textbf{X}= (X_{1}, \cdots, X_{m}) \sim (M, \mu)$, 
$M$ be ergodic with stationary distribution $\pi$.
Then an estimator $\widehat{M}: \Omega^{m} \rightarrow \mathbb{M}_{\Omega}$ exists such that whenever 
\begin{align*}
    m \geq c \max \{ \dfrac{1}{\epsilon^{2}\pi_{*}}\max\{d, \ln{\dfrac{1}{\epsilon \delta_{p}}}\},  \dfrac{1}{\gamma_{ps} \pi_{*}}\ln(\dfrac{d \| \mu/\pi\|_{2, \pi}}{\delta_{p}})\}
\end{align*}
we have, with probability at least $1- \delta_{p}$, 

\begin{align*}
    \|M - \widehat{M}\|_{\infty} < \epsilon, 
\end{align*}
where c is a universal constant, $d= |\Omega|$, $\gamma_{ps}$ is the pseudo-spectral gap, $\pi_{*}$ the minimum stationary probability, and $\| \mu/ \pi\|^{2}_{2,\pi} \leq 1/\pi_{*}$ 
Where these expressions are defined as:
\begin{itemize}
    \item $\gamma_{ps}= \max_{k \geq 1} \{\gamma((M^{*})^{k}M^{k})/k\}$
    \item $\pi_{*}= \min_{i \in \Omega} \pi(i)$
    \item $\| \mu/ \pi\|^{2}_{2,\pi} = \sum_{i \in \Omega} \mu(i)^{2}/\pi(i) \in [1, \infty]$
\end{itemize}
\end{theorem}

\subsection{Approximation and Perturbation Analysis}
In this section we introduce some intermediary results which'll later be required for proving convergence. More specifically we'll convey two things, a. Cost and correlation matrix can be well approximated by finite time horizons, b. Perturbating the policy slightly doesn’t change the cost-value and cost-policy-gradient by a lot.

\begin{restatable}{lem}{costfinite}\label{lemma:costfinite}
For any K with finite C(K), \\
let $C^{l}(K)=\mathbbm{E}_{\omega(0) \sim \pi, x_{0} \sim \mathcal{D}} \left[ \sum_{t=0}^{l-1}x_t^\top Q_{\omega(t)}x_t + u_t^\top R_{\omega(t)}u_t \right]$, if
\begin{equation*}
    l \ge \frac{d\cdot C^2(K)*(\sum_{i=1}^{N_{s}} \|Q_{i}\|+\|R_{i}\|\|K_{i}\|^2)}{\epsilon\mu \Lambda_{min}^2(Q)} 
\end{equation*}
then: 
\begin{equation*}
    C(K) - C^{l}(K)  \le \epsilon 
\end{equation*}
\end{restatable}


For proof, refer Appendix. From Lemma \ref{lemma:costfinite} we can surmise that the roll-out length $l$ has a lower bound above which the accrued cost if $\epsilon$ close to the cost where our policy is rolled-out till infinity. This helps us set an appropriate initialization for the roll-out length $l$, as we'll see later. Next, we introduce an analogous result pertaining to the state correlation matrix $\chi$.

\begin{restatable}{lem}{chiKperturbation}\label{lemma:chiKperturbation}
For any K with finite C(K) \\
let $\chi_{K}^{l}=$ diag$\left( \sum_{t=0}^{l}X_{1}(t), \cdots, \sum_{t=0}^{l}X_{N_{s}}(t)\right)$, where 
$X_{i}(t):=\mathbb{E}[x_{t}x_{t}^{\top}\mathbbm{1}_{\omega(t)=i}]$, now if

\begin{equation*}
    l \ge \frac{d\cdot C^2(K)}{\epsilon\mu \Lambda_{min}^2(Q)}
\end{equation*}
then \\ 
\begin{equation*}
    \|\chi^{(l)}_K - \chi_K\| \le \epsilon
\end{equation*}
\end{restatable}






Akin to Lemma \ref{lemma:costfinite}, here get an idea about the roll-out length required such that the correlation matrices can be well approximated without the need of rolling the simulation to infinity and thus saving time. \\

Next we'll elaborate on perturbation results which are conceptually necessary for proving convergence. It was argued in \cite{Porto et al.} that the almost smoothness requirement, which expresses $\Delta C(K)$ in terms of $\Delta K$, from \cite{Kakade et al.} can't be trivially extended to MJLS's case and provide novel Lyapunov argument to that end. Hinging on that, now introduce lemmas catering to policy perturbation.

\begin{restatable}{lem}{CKperturb}\label{lemma:CKperturb}
Suppose $K'$ is such that: \\
\begin{align*}
    \|K'-K\| \le & \min (\frac{\Lambda_{min}(Q) \mu}{4 C(K) \sum_{i}^{N_{s}} \|B_{i}\|(\|A_{i}-B_{i}K_{i}\|+1)}, \\ & \sum_{i}^{N_{s}}\|K_{i}\|)
\end{align*}
then:
\begin{equation*}
    |C(K')-C(K)| \le c_{diff}(\|K-K'\|)
\end{equation*}
Where $c_{diff}$ is a function of system variables or
\begin{equation*}
    c_{diff}=f\left(\|A\|, \|B\|, \|Q\|, \|R\|, \|K\|, \dfrac{C(K)}{\Lambda_{\min}(Q)}\right)
\end{equation*}
\end{restatable}





Above Lemma \ref{lemma:CKperturb} makes sure that the region around a stable policy $K$ is also stable, something we'll use while concluding convergence in coming sections. Proof can be referred to in the Appendix. The lemma can also be thought of as a sanity check for policy iterations, as we perturb policy along the direction that decreases cost.

\begin{restatable}{lem}{delCKperturb}\label{lemma:delCKperturb}
Suppose $K^{'}$ is such that:\\
\begin{equation*}
    \|K'-K\| \le \min\left(\frac{\Lambda_{min}(Q) \mu}{4 C(K) \sum_{i}^{N_{s}} \|B_{i}\|(\|A_{i}-B_{i}K_{i}\|+1)}, \sum_{i}^{N_{s}}\|K_{i}\|\right)
\end{equation*}

then there is a polynomial $g_{diff}$, a polynomial of system variables\\
such that:\\
\begin{equation*}
\|\nabla C(K')-\nabla C(K)\|
\leq  g_{diff} \|K'-K\|.
\end{equation*}
\end{restatable}
Lemma \ref{lemma:delCKperturb} intuitively says that for a stable policy $K$ the region around won't blow up the cost, since the difference in gradients is bounded, which is something we'll need while showing that gradient estimates $\nabla \widehat{C(K)}$, based on zeroth-order optimization \cite{Flaxman et al.}, are close to the actual gradients. Now that we have some idea about how the landscape of $C(K)$ behaves as we move slightly from stabilizing policies, we'll progress further to proving our main convergence results.

\subsection{Convergence for Algorithm-1}
In this section we introduce the policy convergence result for Algorithm-\ref{model_free_MJLS}
when we use natural policy gradient \cite{Kakade NPG et al.}, and prove that we can converge to optimal policy with lower bounds on both sample and time being polynomial in relevant system variables. Exact proof is shifted to the Appendix which uses several intermediary results but we do give an outline of the proof below. 

\begin{theorem}[Main Result]\label{thm:NGDforMJLS}




Suppose $C(\hat{K}_0)$ is finite and $\mu \ge 0$, and that the gradient $\nabla C(\hat{K}_{t})$ and state correlation matrix $\chi_{K}$ are estimated via Algorithm-1 then for different policy update rules with corresponding conditions such that, \\\\ 
\textbf{For gradient descent:}\\
The update rule
$$
\hat{K}_{t+1} = \hat{K}_t - \eta \nabla C(\hat{K}_t)
$$
with step size
$$
\eta \le \frac{1}{eta_{grad}}
$$
and
$$r = 1/f_{GD, r}(1/\epsilon)$$ along with $$m \ge f_{GD, sample}(d,
1/\epsilon, L^2/\mu)$$ samples, both are truncated to
$f_{GD,\it}(d,1/\epsilon)$ iterations, then with high probability
(at least $1-\exp(-d)$) in $T$ iterations where
\[
T>\frac{\|X^{K^*} \|}{\mu \Lambda_{\textrm{min}}(R) } \
\left({eta_{grad}}\right)\log \frac{2(C(\hat{K_0}) -C(\hat{K^*}))}{3 \eps}
\] Where the $eta_{grad}$ is defined in the  equation \ref{eq:etagrad} in the appendix.\\
The gradient descent update rule satisfies:\\
$$
C(\hat{K}_T) - C(\hat{K}^*) \le \epsilon
$$
\\
\textbf{For natural gradient descent:}\\
With update rule
$$
\hat{K}_{t+1} = \hat{K}_t - \eta \nabla C(\hat{K}_t) (\chi^{\hat{K}_t})^{-1}
$$
with step size
$$
\eta \le \frac{1}{2\left( \| {R} \| + \frac{\| {B} \|^2 C({K}_0)}{\mu} \right)}
$$
and 
$$r = 1/f_{NGD, r}(1/\epsilon)$$ along with $$m \ge f_{NGD, sample}(d,
1/\epsilon, L^2/\mu)$$ samples, both are truncated to
$f_{NGD,\it}(d,1/\epsilon)$ iterations, then with high probability
(at least $1-\exp(-d)$) in $T$ iterations where
\[
T>\frac{\|X^{K^*} \|}{\mu} \,
\left(\frac{\|R\|}{\Lambda_{\textrm{min}}(R)} + 
\frac{\|B\|^2 C(\hat{K_0})}{\mu \Lambda_{\textrm{min}}(R)} \right) 
\, \log \frac{2(C(\hat{K_0}) -C(\hat{K^*}))}{3 \eps}
\]
The natural gradient descent update rule satisfies:\\
$$
C(\hat{K}_T) - C(\hat{K}^*) \le \epsilon
$$
\end{theorem}
\begin{proof}
We obtain the proof by combining results of Theorem-3 and Theorem-4 from the appendix.
\end{proof}

Proofs for Theorem-3 and Theorem-4, which result in Theorem-1, can be found in the appendix below, the schema of the proofs involves intermediary steps showing us that with large enough rollout length $l$, for a given policy, the cost $C(K)$ and state-correlation matrix $\chi$ are $\epsilon$ close to the case when the rollout length is infinity. Next, we show that our estimated $\nabla C(K)$ and $\chi$ can be $\epsilon$ close to the exact gradient and state correlation matrix with enough samples -- which is then used to prove one step bound, which is then extended to prove the above main result. 
\section{Experiments}
In this section we have preformed the experiments. We have implemented Model free Gradient descent and Natural Gradient Descent and compared their convergence with Exact (Model Based) Gradient Descent and Natural Gradient Descent. \par
We have performed the experiments for 2,4 and 6 number of states and same number of action and state dimension as states. \par
We have chosen A and B to be random and done scaling of matrices such that $\lambda_{max}(A) \leq 1$, to keep system stable. Also for the system we have ensure that the $C(K_0)$ is finite where $K_0 = 0$ for all states. Probability of Markov chain obtained using Numpy Dirichlet function. 

These are the results obtained. Below figures shows the variation of Normalized cost difference with number of iteration of update of policy.
\begin{figure}[!htb]
    \centering  
    \includegraphics[width=0.7\textwidth]{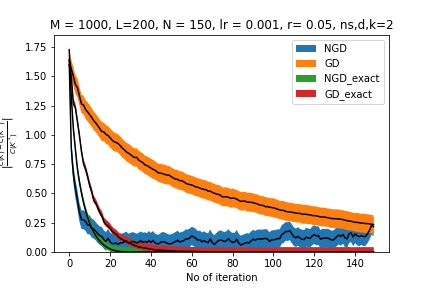}
    \caption{Results for 2 states, 2 inputs, 2 modes}
    \label{fig:States_2}
\end{figure}

\begin{figure}[!htb]
    \centering
    \includegraphics[width=0.7\textwidth]{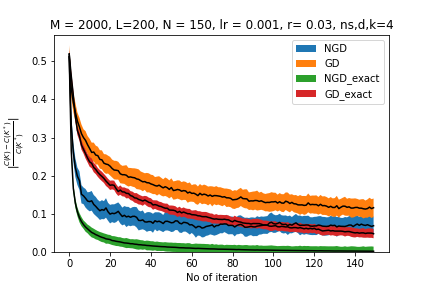}
    \caption{Results for 4 states, 4 inputs, 4 modes}
    \label{fig:States_4}
\end{figure}

\begin{figure}[!htb]
    \centering
    \includegraphics[width=0.7\textwidth]{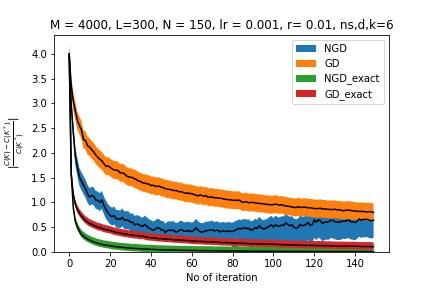}
    \caption{Results for 6 states, 6 inputs, 6 modes}
    \label{fig:States_6}
\end{figure}

\textbf{Remarks}
\begin{itemize}
    \item From the plots we can see that Model free Natural Gradient Descent performs approximately similar to the case of Model Based Gradient Descent(A,B are known)
    \item As the Number of states increased the selection of hyper-parameters become difficult and with not so good hyper parameters the algorithm may not even converge
    \item As number of states increases the Difference between the Model Bases NGD and Model Free NGD increases
    \item Roll out length, Number of Trajectory and value of r (smoothing parameters) should be chosen very selectively for convergence
    \item We have performed the every experiment 15 times and taken average to show smooth convergence
\end{itemize}
\section{Conclusion}
We've thus shown policy convergence both theoretically and simulation wise, and thus approached the open problem mentioned in \cite{Porto et al.}. We envision our work to provide basis for real world applications that can be modelled using MJLS where system parameters are unknown, and the optimal policy needs to be arrived at by gathering indirect information. Furthermore, it needs to be explored how policy convergence would occur in cases where we don't observe states directly or where system changes continuously instead of changing in discrete time steps. We keep these questions for further study.
\section{Appendix}

\costfinite*
\begin{proof}
define: $$ P^K = (P_1, \ldots, P_{n_s})$$
where $P_i$ is defined as in equation \eqref{eq:6} \\
Now cost function can be written as :\\
\[
C(K) = \Expxt{x_0^T \left( \sum_{i\in\Omega } \rho_i P_i^{K} \right) x_0}
\]
Hence C(K) can be written as: \\
\begin{align*}
C(K) &= \langle P^K , X(0) \rangle \\
&=  \langle Q+K^\top RK, \cT(X(0) \rangle \\
&=  \langle Q+K^\top RK, \chi_K \rangle \\
&= \sum_{i=1}^{n_s} tr((Q_i+K^\top_i R_i K_i)\chi_{K,i}) \\
& \ge \Lambda_{min}(Q)\ trace(\chi_K) \\
So:\\
trace(\chi_K) & \le \frac{C(K)}{\Lambda_{min}(Q)}
\end{align*}

Now consider: \\
\begin{align*}
   & \sum_{i=0}^{\it-1}trace(\cF^i(X(0)))\\
    &= trace(\sum_{i=0}^{\it-1} \cF^i(X(0))) \\
    & \le trace(\sum_{i=0}^{\infty} \cF^i(X(0))) \\
    &= trace(\cT_{K}(X(0)))\\
    &= trace(\chi_K) \\
    & \le \frac{d C(K)}{\Lambda_{min}(Q)} \\
\end{align*}
Since all traces are nonnegative, then there must exist $j \le \it $ such that: \\

$$trace(\chi_{K}^{j})  \le \frac{d.C(K)}{ \Lambda_{min}(Q) \it}$$ \\
also: $$ \chi_K \preceq \frac{C(K)*X(0)}{\mu \Lambda_{min}(Q)} $$ \\
and 
\begin{align*}
trace(\cF^j(\chi_K)) &\le \frac{C(K)*trace(\cF^j(X(0)))}{\mu \Lambda_{min}(Q)}\\ &\le \frac{d.C^2(K)}{l \mu \Lambda_{min}^2(Q)}
\end{align*}
so as long as $ l \ge \frac{d.C^2(K)}{\eps \mu \Lambda_{min}^2(Q)}$ \\
$\|\chi^{(\it)}_K - \chi_K\| \le \|\chi^{(j)}_K - \chi_K\| = \|\cF^j(\chi_K) \| \le trace(\cF^j(\chi_K)) \le \eps $ \\

\end{proof}

\chiKperturbation*
\begin{proof}
we know: \\
$$C(K) = \langle \chi_K,Q+K^\top RK \rangle$$
$$C^\it(K) = \langle \chi^\it_K,Q+K^\top RK \rangle$$
Now:\\
$$C(K)-C^\it(K) \le trace(\chi_K-\chi^\it_K)(\|Q\|+\|R\|\|K\|^2) $$
therefore if :\\
$$ \it \ge \frac{d\cdot C^2(K)*(\|Q\|+\|R\|\|K\|^2)}{\epsilon\mu \Lambda_{min}^2(Q)}$$
then: $$ trace(\chi_K-\chi^\it_K) \le \frac{\eps}{\|Q\|+\|R\|\|K\|^2} $$ above is from lemma 1. \\
Hence: $$ C(K) - C^\it(K)  \le \eps $$
\end{proof}

\CKperturb* 
\begin{proof}
Let $L_{i}^{K}= (R_{i}+B_{i}^{\top}\mathcal{E}_{i}(P^{K})B_{i})K_{i}- B_{i}^{\top}\mathcal{E}_{i}(P^{K})A_{i}$, and $\psi= R+B^{\top}\mathcal{E}(P^{K})B$. And so we can write, 

\begin{align}\label{eq:14}
C(K^{'})-C(K)= -2\langle \Delta K^{\top}L^{K}, \chi_{K^{'}}\rangle+\langle\Delta K^{\top}\psi \Delta K, \chi_{K^{'}} \rangle
\end{align}
Thus expanding equation (\ref{eq:14}) we get, 
\begin{align*}
\begin{split}
    C(K^{'})-C(K){}&= 2\langle\Delta K^{\top}[B_{i}^{\top}\mathcal{E}_{i}(P^{K})A_{i}-(R_{i}+B_{i}^{\top}\mathcal{E}_{i}(P^{K})B_{i})K_{i}], \\
    &{}\chi_{K^{'}} \rangle +
\langle \Delta K^{\top} (R+B^{\top}\mathcal{E}(P^{K})B) \Delta K, \chi_{K^{'}}\rangle\\
\end{split}
\end{align*}

We know that 
\begin{align*}
\langle A,B \rangle &= Tr(A^{\top}B) \\
Tr(ABC) &= Tr(CAB) \\
    Tr(ABC) &\leq Tr(AB)Tr(C) \\
    &\leq  Tr(AB)\{d\times \|C\|\} \cdots \text{d: dimension of C}
\end{align*}
Let $\Delta C(K)= C(K^{'})-C(K)$, therefore we can write, 
\begin{align*}
    \Delta C(&K)= 2\sum_{i= 1}^{N_{s}} Tr\left( [B_{i}\mathcal{E}(P_{i}^{K})A_{i}- (R_{i}B_{i}^{\top}\mathcal{E}(P_{i}^{K})B_{i})K_{i}]^{\top}\Delta K_{i}X_{i}^{K^{'}}\right) \\
    &+ \sum_{i= 1}^{N_{s}}Tr\left( \Delta K_{i}^{\top}(R_{i}+B_{i}^{\top}\mathcal{E}(P_{i}^{K}B_{i})^{\top})\Delta K_{i}X_{i}^{K^{'}}\right) \\
    &\leq 2 \sum_{i= 1}^{N_{s}}(\|\Delta K_{i}\|.\|B_{i}\mathcal{C}(P_{i}^{K})A_{i}-(R_{i}B_{i}^{\top}\mathcal{E}(P_{i}^{K})B_{i})K_{i}\|.Tr(X_{i}^{K^{'}})) \\
    &+\sum_{i= 1}^{N_{s}}\|\Delta K_{i}\|^{2}.\|B_{i}\mathcal{E}(P^{K}_{i}B_{i})+R_{i}\| \times Tr(X_{i}^{K^{'}}) \\
    & \leq 2\|\Delta K\|_{\max}.\left(\|B\|_{\max}\|A\|_{\max}\|P^{K}\|_{\max}+\|R\|_{\max}.\|B\|_{\max}.\|P^{K}\|_{\max}.\|K\|_{\max}\right) \times (\sum_{i= 1}^{N_{s}}Tr(X_{i}^{K^{'}}))\\
    &+\|\Delta K\|_{\max}^{2}.\left(\|B\|_{\max}^{2}\|P^{K}\|_{\max}+\|R\|_{\max}\right)\times(\sum_{i= 1}^{N_{s}} Tr(X_{i}^{K^{'}}))
\end{align*}
We can also write,
\begin{align*}
    \|\Delta K\|_{\max}^{2}&= \|K-K^{'}\|_{\max}^{2} \\
    &=\|K-K^{'}\|_{\max}\times\|K-K^{'}\|_{\max} \\
    &\leq \|K\|_{\max}\|K-K^{'}\|_{\max}
\end{align*}
Hence we get, 

\begin{align*}
    \Delta C(K) \leq & h_{diff}^{1}.\|\Delta K\|_{\max}.\left( \sum_{i\in\Omega}Tr(X_{i}^{K^{'}})\right) \\
    & h_{diff}^{2}.\|\Delta K\|_{\max}.\left(\sum_{i= 1}^{N_{s}}Tr(X_{i}^{K^{'}})\right)
\end{align*}
 
 Where $h_{diff}^{1}, h_{diff}^{2} \sim f\left( \|B\|, \|A\|, \|P^{K}\|, \|K\|, \|R\|\right)$
 \begin{align}\label{eq:15}
     \implies C(K^{'})-C(K) \leq (h_{diff}^{1}+h_{diff}^{2}).\|\Delta K\|_{\max}.(\sum_{i= 1}^{N_{s}}Tr(X_{i}^{K^{'}}))
 \end{align}
 from A.3 in \cite{Porto et al.} we get, $\sum_{i= 1}^{N_{s}}Tr(X_{i}^{K}) \leq \dfrac{C(K)}{\Lambda_{\min}(Q)}$. For gradient/natural gradient descent $K \rightarrow K^{'}$ updates we know: $C(K^{'}) < C(K)$. Thus we get, 
 \begin{align}\label{eq:16}
     \sum_{i= 1}^{N_{s}}Tr(X_{i}^{K^{'}}) \leq \dfrac{C(K^{'})}{\Lambda_{\min}(Q)} \leq \dfrac{C(K)}{\Lambda_{\min}(Q)}
 \end{align}
 
 Using equations (\ref{eq:14} and \ref{eq:15})we get,
 
 \begin{align*}
     |C(K^{'})-C(K)| \leq h^{4}_{diff}.\|K-K^{'}\|_{\max} 
 \end{align*}
 
 where $h_{diff}^{4} \sim f(\|A\|, \|B\|, \|Q\|, \|R\|, \|K\|, \dfrac{C(K)}{\Lambda_{\min}(Q)})$ or 
 \begin{align*}
 \begin{split}
 h_{diff}^{4}&= \dfrac{C(K)}{\Lambda_{\min}(Q)}. (2.\|B\|_{\max}\dfrac{C(K)}{\mu}(\|A\|_{\max}\\ &+\|R\|_{\max}\|K\|_{\max})+ \|B\|_{\max}^{2}\dfrac{C(K)}{\mu}\\
 &+\|R\|_{\max} ) \cdots \cdots \text{ using sideresult from Theorem 3 in \cite{Porto et al.}}
  \end{split}
 \end{align*}
\end{proof}
\delCKperturb*
\begin{proof}
We know that $\nabla C(K)= 2L^{K}\chi^{K}$ where, $L_{i}^{K}= (R_{i}+B_{i}^{\top}\mathcal{E}_{i}(P^{K})B_{i})K_{i}- B_{i}^{\top}\mathcal{E}_{i}(P^{K})A_{i}$\\
We can write, 
\begin{align*}
    \nabla C(K^{'})-\nabla &C(K)=2(L^{K^{'}}\chi^{K^{'}}- L^{K}\chi^{K})\\
    &= 2(L^{K^{'}}\chi^{K^{'}}-L^{K}\chi^{K^{'}}+L^{K}\chi^{K^{'}}-L^{K}\chi^{K})\\
    &= 2((L^{K^{'}}-L^{K})\chi^{K^{'}}+L^{K}(\chi^{K^{'}}-\chi^{K}))
\end{align*}
We now have 4 terms to bound:
\begin{itemize}
    \item $L^{K^{'}}-L^{K}$
    \item $\chi^{K^{'}}$
    \item $L^{K}$
    \item $(\chi^{K^{'}}-\chi^{K})$
\end{itemize}

Now from equation A.3 in the proof of Lemma 3 from \cite{Porto et al.} that 
\begin{align*}
    \sum_{i= 1}^{N_{s}}tr(X_{i}^{K^{'}}) \leq \dfrac{C(K^{'})}{\Lambda_{\min}(Q)} \leq \dfrac{C(K)}{\Lambda_{\min}(Q)}\\
\end{align*}
\begin{equation}\label{eq:17}
    \implies \|\chi_{K^{'}}\| \leq \dfrac{C(K)}{\Lambda_{\min}(Q)}
\end{equation}

Bounding $L^{K}$ is relatively straightforward. We know that, 
\begin{align*}
    C(K)= \mathbb{E}_{x_{0} \sim D}[tr((\sum_{i= 1}^{N_{s}} \pi_{i}P_{i}^{K})x_{0}x_{0}^{\top})]\\
    \implies \|P^{K}\|_{\max} \leq \dfrac{C(K)}{\mu}
\end{align*}

Using the above inequality we can thus bound $L^{K}$ as, 
\begin{align*}
    \|L^{K}\|&= \sum_{i= 1}^{N_{s}}\|(R_{i}+B_{i}^{\top}\mathcal{E}_{i}(P^{K})B_{i})K_{i}- B_{i}^{\top}\mathcal{E}_{i}(P^{K})A_{i}\| \\
    & \leq \sum_{i= 1}^{N_{s}}\|(R_{i}+B_{i}^{\top}\mathcal{E}_{i}(P^{K})B_{i})K_{i}+ B_{i}^{\top}\mathcal{E}_{i}(P^{K})A_{i}\| \\
    & \leq N_{s}((\|R\|_{\max}+\|B\|_{\max}\dfrac{C(K)}{\mu})\|K\|_{\max}\\
    &+\|B\|_{\max}\|A\|_{\max}\dfrac{C(K)}{\mu})
\end{align*}

Thus, using results from Lemma \ref{lemma:delLk}, \ref{lemma:delchik} and the above bounds for $\chi_{K^{'}}$ and $L^{K}$ we get that $\|\nabla C(K^{'})-\nabla C(K)\| \leq g_{diff}\|K^{'}-K\|$ where $g_{diff} \sim f(\|A\|, \|B\|, \|Q\|, \|R\|, \|K\|)$

\end{proof}

\begin{restatable}{lem}{delchik}\label{lemma:delchik}
Suppose $K^{'}$ is such that:\\
\begin{equation*}
    \|K'-K\| \le \min\left(\frac{\Lambda_{min}(Q) \mu}{4 C(K) \sum_{i}^{N_{s}} \|B_{i}\|(\|A_{i}-B_{i}K_{i}\|+1)}, \sum_{i}^{N_{s}}\|K_{i}\|\right)
\end{equation*}

then there is a polynomial $g_{\chi diff}$\\
such that:\\
\begin{equation*}
\| \chi^{K^{'}} - \chi^{K}\|
\leq  g_{\chi, diff} \|K'-K\|.
\end{equation*}
\end{restatable}

\begin{proof}
Let us define $\bar{x}_{i,t}$ to be the state matrix evolving with time when the system is LQR, $p_{ij}=1$ if $i=j$ else 0, with state dynamics $A_{i}, B_{i}, Q_{i}, R_{i}$ instead of MJLS with same initial state $x_{0}$.
\begin{align*}
    X_{i}^{K}= \sum_{t=0}^{\infty}X_{i}(t)&= \sum_{t=0}^{\infty}\mathbb{E}[x_{t}x_{t}^{\top}\mathbbm{1}_{\omega(t)=i}]\\
    \|\chi_{K^{'}} -\chi_{K}\|&= \sum_{i= 1}^{N_{s}} \|X_{i}^{K^{'}}- X_{i}^{K}\|
\end{align*}

We can write, 
\begin{align*}
    \chi_{K^{'}_{i}}- \chi_{K_{i}}&= \sum_{t=0}^{\infty}\mathbb{E}[(x_{t}^{K^{'}}x_{t}^{K^{'}}- x_{t}^{K}x_{t}^{K}).\mathbbm{1}_{\omega(t)=i}] \\
    &\leq \max_{i}\{\sum_{t=0}^{\infty}\mathbb{E}[(\bar{x}_{i,t}^{K^{'}}\bar{x}_{i,t}^{K^{'}}- \bar{x}_{i,t}^{K}\bar{x}_{i,t}^{K})]\}
\end{align*}
Now, we know from Lemma 16 in \cite{Kakade et al.} that if $K^{'}-K \leq \dfrac{\sigma_{\min}(Q)\mu}{4C(K)\|B\|(\|A-BK\|+1)}$ it holds that 
\begin{align*}
    \|\Sigma_{K^{'}}-\Sigma_{K}\| \leq 4(\dfrac{C(K)}{\sigma_{\min}(Q)})^{2}\dfrac{\|B\|(\|A-BK\|+1)\|K-K^{'}\|}{\mu}
\end{align*}
Using the above result we can write 
\begin{align*}
    \|X_{i}^{K^{'}}-&X_{i}^{K}\| \leq\\ &\max_{i}\{4(\dfrac{C_{LQR}(K)^{i}}{\sigma_{\min}(Q_{i})})^{2}\dfrac{\|B\|_{i}(\|A_{i}-B_{i}K_{i}+1\|)\|K_{i}-K_{i}^{'}\|}{\mu}\}
\end{align*}

Where $C_{LQR}(K)^{i}$ is the cost accrued were the system LQR with state dynamics $A_{i}, B_{i}, Q_{i}, R_{i}$ instead of MJLS with same initial state $x_{0}$.\\\\

We can thus write,
\begin{align*}
    \|& \chi_{K^{'}}-\chi_{K}\| = \sum_{i= 1}^{N_{s}} \|X_{i}^{K^{'}}-X_{i}^{K}\| \\
    &\leq N^{2}_{s}.\max_{i}\{4(\dfrac{C_{LQR}(K)^{i}}{\sigma_{\min}(Q_{i})})^{2}\dfrac{\|B\|_{i}(\|A_{i}-B_{i}K_{i}+1\|)}{\mu}\}.\|K-K^{'}\|
\end{align*}
\end{proof}

\begin{restatable}{lem}{delLk}\label{lemma:delLk}
Suppose $K^{'}$ is such that:\\
\begin{equation*}
    \|K'-K\| \le \min\left(\frac{\Lambda_{min}(Q) \mu}{4 C(K) \sum_{i}^{N_{s}} \|B_{i}\|(\|A_{i}-B_{i}K_{i}\|+1)}, \sum_{i}^{N_{s}}\|K_{i}\|\right)
\end{equation*}

then there is a polynomial $g_{\chi diff}$ \\
such that:\\
\begin{equation*}
\| L^{K^{'}} - L^{K}\|
\leq  g_{L, diff} \|K'-K\|.
\end{equation*}
\end{restatable}

\begin{proof}
We know that, 
\begin{align*}
    L_{i}^{K}= (R_{i}+B_{i}^{\top}\mathcal{E}_{i}(P^{K})B_{i})K_{i}-B_{i}^{\top}\mathcal{E}_{i}(P^{K})A_{i}
\end{align*}
also, $\|L^{K^{'}}- L^{K}\| = \sum_{i= 1}^{N_{s}} \|L_{i}^{K^{'}}-L_{i}^{K}\|$ thus we get,
\begin{align*}
    L_{i}^{K^{'}}-L_{i}^{K}&= R_{i}(K^{'}_{i}-K_{i})- B_{i}^{\top}(\mathcal{E}_{i}(P^{K^{'}}- P^{K}))A_{i}\\
    &+B_{i}^{\top}(\mathcal{E}_{i}(P^{K^{'}}-P^{K})).B_{i}K_{i}^{'}\\
    &+B_{i}\mathcal{E}_{i}(P^{K})B_{i}(K^{'}_{i}-K_{i})
\end{align*}

We thus need to bound $\mathcal{E}_{i}(P^{K^{'}}-P^{K})$ which we'll do next.\\\\
We know that $C(K)= \mathbb{E}_{x_{0} \sim D}[tr((\sum_{i= 1}^{N_{s}},  \pi_{i}P^{K}_{i})x_{0}x_{0}^{\top})]$. Let $\Delta C(K)= C(K^{'})-C(K)$, we then get:
\begin{align*}
    \Delta C(K)&= \mathbb{E}_{x_{0} \sim D}[tr((\sum_{i= 1}^{N_{s}} \pi_{i}(P^{K^{'}}_{i}-P^{K}_{i}))x_{0}x_{0}^{\top})] \\
    &\geq tr(\sum_{i= 1}^{N_{s}}\pi_{i}(P^{K^{'}}_{i}-P_{i}^{K})).\sigma_{\min}(\mathbb{E}_{x_{0}\sim D}(x_{0}x_{0}^{\top}))\\
    &\geq tr(\sum_{i= 1}^{N_{s}}(P^{K^{'}}-P_{i}^{K})).\min(\pi_{i}).\sigma_{\min}(\mathbb{E}_{x_{0}\sim D}(x_{0}x_{0}^{\top})\\
    \implies &tr(\sum_{i= 1}^{N_{s}}(P^{K^{'}}-P_{i}^{K})) \leq \dfrac{|C(K^{'})-C(K)|}{\mu}
\end{align*}

From Lemma \ref{lemma:delCKperturb} we know that $|C(K^{'})-C(K)| \leq h_{diff}\|K^{'}-K\|$
\begin{align*}
    \|\sum_{i= 1}^{N_{s}}(P_{i}^{K^{'}}-P_{i}^{K})\| \leq h_{diff}.\dfrac{(\|K^{'}-K\|)}{\mu}
\end{align*}
Now, 
\begin{align*}
    \mathcal{E}_{i}(P^{K^{'}}-P^{K})&= \sum p_{ij}(P^{K^{'}}_{j}-P_{j}^{K})\\
    \|\mathcal{E}_{i}(P^{K^{'}} - P^{K})\|&= \|\sum p_{ij}(P_{j}^{K^{'}}- P_{j}^{K})\|\\
\end{align*}
\begin{equation}\label{eq:18}
    \|\mathcal{E}_{i}(P^{K^{'}} - P^{K})\| \leq h_{diff}.\dfrac{(\|K^{'}-K\|)}{\mu}
\end{equation}
Substituting  equation \ref{eq:17} in the expression for $L_{i}^{K^{K}}-L_{i}^{K}$ above we get, 
\begin{align*}
    \|L^{K}- L^{K^{'}}\| &\leq \|R\|_{\max}\Delta K+ \|A\|_{\max}\|B\|_{\max}\dfrac{h_{diff}\Delta K}{\mu}\\
    &+\|B\|_{\max}\dfrac{h_{diff}\Delta K}{\mu}\|K\| \\
    &+\|B\|^{2}_{\max}\dfrac{C(K) \Delta K}{\mu}
\end{align*}

Thus $\|L^(K)-L^{K^{'}}\|= h_{L}\Delta K$ where $h_{L} \sim f(\|A\|, \|B\|, \|K\|, C(K_{0}), h_{diff})$.
\end{proof}
In the lemma \ref{lemma:equivalence_tp} below we show why it makes sense to use the estimated transition probability matrix in case we don't know them beforehand. Owing to closeness shown between estimated gradients and state correlation matrix in the cases where the transition probabilities are known and estimated, with high probability, in the convergence proofs we can use the terms meant for the case when the transition probability's known for lucidity's sake.

\begin{restatable}{lem}{equivalence_tp}\label{lemma:equivalence_tp}
Let $\nabla \widehat{C}(K)$ and $\hat{\chi}_{K}$ be the estimated gradient and expected state correlation matrix when the transition probabilities are exactly known, and correspondingly let $\nabla \widehat{C}(K)_{est}$ and $\hat{\chi}_{est, K}$ be the value when the transition probabilities are estimated to a large precision ($\|P-\hat{P} \|_{\infty} < \epsilon_{p}$) with a high probability, then for small enough $\epsilon_{p}$,  $\exists \epsilon_{\nabla}, \epsilon_{\chi}$ such that $\| \nabla \widehat{C}(K)- \nabla \widehat{C}(K)_{est}\| \leq \epsilon_{\nabla}$ and $\|\hat{\chi}_{K}- \hat{\chi}_{est,K}\| \leq \epsilon_{\hat{\chi}}$ with high probability.
\end{restatable}
\begin{proof}
The lemma is proved in two parts below, we first prove the error bound between expected state correlation matrix $\hat{\chi}_{K}$, $\hat{\chi}_{est, K}$ and then correspondingly use the result to prove bound on gradients. The proof exploits the recursive nature of the correlation matrix, with trying to bound the difference between two terms at each time step. Furthermore this bound on expected difference is then used to bound the actual difference using hoeffding's inequality.\\

Part a: \\\\
We know that $X_{i}(t):=\mathbb{E}[x_{t}x_{t}^{\top}\mathbbm{1}_{\omega(t)=i}]$, and 
\begin{equation*}
    X_{j}(t+1)= \sum_{i= 1}^{N_{s}} p_{ij}(A_{i}-B_{i}K_{i})X_{i}(t)(A_{i}-B_{i}K_{i})^{\top}
\end{equation*}

Now, for a policy $K$, let $\hat{X_{j}}(t+1)$ be the matrix when the estimated transition probability matrices, $[\hat{p}_{ij}]$, are used.

\begin{align*}
     X_{j}(t+1)-\hat{X_{j}}(t+1)=\sum_{i= 1}^{N_{s}}(A_{i}-B_{i}K_{i})&\left[p_{ij}X_{i}(t)-\hat{p}_{ij}\hat{X_{i}}(t)\right]\\
     & \times(A_{i}-B_{i}K_{i})^{\top}
\end{align*}

now, \textit{assume} that $\|X_{i}(t)- \hat{X}_{i}(t)\| \leq \epsilon_{t} \cdots \cdots w.h.p$ \\
The assumption primarily serves to observe how the bound varies with time $t$. \\\\
We can then write, 

\begin{align*}
    p_{ij}X_{i}(t)- \hat{p}_{ij}\hat{X}_{i}(t)= (p_{ij}- \hat{p}_{ij})X_{i}(t)+\hat{p}_{ij}(X_{i}(t)- \hat{X}_{i}(t))
\end{align*}

Since we know that, $\chi_{K} \leq \dfrac{C(K)*X(0)}{\mu \Lambda_{\min}(Q)}$ \\

\begin{align*}
    \implies \|X_{i}(t)\| \leq \epsilon_{K}
\end{align*}

where $\epsilon_{K}$ is an upper bound which is a function of $K$. Thus we get, 

\begin{align*}
    &\|(p_{ij}- \hat{p}_{ij})X_{i}(t)+\hat{p}_{ij}(X_{i}(t)- \hat{X}_{i}(t))\| \leq \epsilon_{p}\epsilon_{K}+\epsilon_{t} 
\end{align*}

\begin{equation} \label{eq:19}
    \implies \|X_{i}(t+1)- \hat{X}_{j}(t+1)\| \leq N_{s}c_{K}(\epsilon_{p}\epsilon_{K}+\epsilon_{t})
\end{equation}

Where $c_{K}= N_{s}(\|A\|_{\max}+\|B\|_{\max}\|K\|_{\max})$ is a constant. Now that we have a recursive relationship over bound, we can exploit bound from $t=1$ to get the remaining ones. \\\\
For $t=1$, 
\begin{align*}
    & X_{j}(1)- \hat{X}_{j}(1) \leq \sum_{i=1}^{N_{s}} (A_{i}-B_{i}K_{i})x_{0}x_{0}^{\top}(A_{i}-B_{i}K_{i})^{\top}*(p_{ij}-\hat{p}_{ij})\\
\end{align*}

\begin{equation}\label{eq:20}
    \|X_{j}(1)- \hat{X}_{j}(1)\| \leq c_{K}^{1}\epsilon_{p}
\end{equation}

Where $c_{K}^{1}= N_{s}(\|A\|_{\max}+ \|B\|_{\max}\|K\|_{\max})^{2}\|x_{0}x_{0}^{\top}\|$ is a constant. Thus the bounds (\ref{eq:14}) can be recursively determined starting from $t=1$ (\ref{eq:15}). We thus get, \\

\begin{equation} \label{eq:21}
    \sum_{t=1}^{T}\left(\mathbb{E}[x_{t}x_{t}^{\top}\mathbbm{1}_{\omega(t)=i}] -\mathbb{E}[\hat{x}_{t}\hat{x}_{t}^{\top}\mathbbm{1}_{\omega(t)=i}] \right) \leq \sum_{t=1}^{T} \epsilon_{t}= \epsilon_{X,T}
\end{equation}

Now from the algorithm we know that, $\widehat \chi_K = \frac{1}{m} \sum_{i=1}^m\textup{diag}\left( \sum_{j=1}^{l}x_{t}x_{t}^{\top}\mathbbm{1}_{\omega(j)=1},\cdots, \sum_{j=1}^{l}x_{t}x_{t}^{\top}\mathbbm{1}_{\omega(j)=N_{s}}\right)$. Let $\chi_{est, K}$ be the correlation matrix using estimated transition probabilities. We thus get, 

\begin{align*}
    \sum_{i=1}^{N_{s}}\mathbb{E}[\chi_{K,i}- \chi_{est, K, i}]= \sum_{i=1}^{N_{s}}\sum_{t=1}^{l} (\mathbb{E}[x_{t}x_{t}^{\top}]- \mathbb{E}[\hat{x}_{t}\hat{x}_{t}^{\top}])\mathbbm{1}_{\omega(t)=i}
\end{align*}

Now using Hoeffding bound, since we average over $m$ trajectories, we know that:

\begin{align*}
    P\left[\| \chi_{K}- \chi_{est, K}- \mathbb{E}[\chi_{K}- \chi_{est, K}]\| \geq \delta \right] \leq \exp^{-2m\delta^{2}}
\end{align*}
Thus using above result and \ref{eq:16} we conclude that $\chi_{K}$ and $\chi_{est, K}$ are $\delta+ \epsilon_{X,T}$ close to each other with a high probability. \\\\

Part b: \\\\
Let $P_i$ be defined as in equation \eqref{eq:6} \\
Now cost function can be written as :\\
\[
C(K) = \Expxt{x_0^T \left( \sum_{i\in\Omega } \rho_i P_i^{K} \right) x_0}
\]
Thus: \\
\begin{align*}
C(K) &= \langle P^K , X(0) \rangle \\
&=  \langle Q+K^\top RK, \cT(X(0) \rangle \\
&=  \langle Q+K^\top RK, \chi_K \rangle \\
&= \sum_{i=1}^{n_s} tr((Q_i+K^\top_i R_i K_i)\chi_{K,i}) \\
\end{align*}

Let $\widehat{C}_{est, i}$ be the cost accumulated, as in Algorithm 1, by due to the $i-th$ trajectory on using estimated transition probabilities. We can thus write, 

\begin{align*}
    \widehat{C}-\widehat{C}_{est}= \sum_{i=1}^{N_{s}}tr((Q_{i}+K_{i}^{\top}R_{i}K_{i})[\chi_{K, i}- \chi_{est, K, i}]) \\
    \widehat{\nabla C(K)}-\widehat{\nabla C(K)_{est}} = \frac{1}{m} \sum_{i=1}^m d (\widehat{C}_{i}- \widehat{C}_{est, i})
\end{align*}

Since we've already proven bounds on $\chi_{K}- \chi_{est, K}$ in (Part a) above, the cost due to estimated transition probabilities $\nabla \widehat{C}_{est}(K)$ can thus trivially shown to be close to $\nabla \widehat{C}(K)$ with high probability.
\end{proof}

\subsection{Analysis of smoothing parameter and gradient calculation}
In this section we will see how the value of smoothing parameters in chosen and how it will affect cost value and gradient values. 
We have to be careful about choosing the value of smoothing parameters as many times the cost $C(K+r)$ might not be defined. 

In the next lemma we will use the standard technique (e.g. in \cite{Flaxman et al.}) to show

We will denote $\mathbb{S}_{r}$  to represent the uniform distribution over the points with norm $r$ and $\mathbb{B}_r$represent the uniform distribution over all points with norm at most $r$ (the entire sphere).
 
Now we will use the below function and then apply the calculate the gradient using the use the standard technique (e.g. in \cite{Flaxman et al.}) to show that the gradient can be just with an oracle for function value.

$$
C_r(K) = \E_{U\sim \mathbb{B}_{r}}[C(K+U)].
$$
$$
\nabla C_r(K) = (\nabla C_r(K)_1, \ldots, \nabla C_r(K)_{n_s}))
$$
We will calculate the gradient using the lemma 2.1 in the \cite{Flaxman et al.}.

\begin{restatable}{lem}{stokes}\label{lemma:stokes}
$\nabla C_r({K}) = \frac{d}{r^2}\E_{U\sim \mathbb{S}_{r}}[C({K}+U)U]$ 
Here: $C({K}+U)U  = (C({K_1}+U)U, \ldots, C(K_{n_s}+U)U)$ \\
\end{restatable}
\begin{proof}
In this we have used the same U rather then $U_i$ for different states. Now $C(K)$ can be seen as $n_s$ variable function and we are estimating the gradient for every variable.
The proof is similar to lemma Lemma 2.1 in \cite{Flaxman et al.}.

By Stokes formula, 
\begin{align*}
\nabla \int_{\delta \mathbb{B}_r} C(K_1+U,& \ldots K_{n_s}+U) dx =\\
& \int_{\delta \mathcal{S}_r} C(K_1+U, \ldots K_{n_s}+U)\frac{[U,\ldots,U]}{\|U\|_F} dx.
\end{align*}

By definition
$$
C_r(K) = \frac{\int_{\delta \mathbb{B}_r} C(K_1+U,\ldots, K_{n_s}+U) dx}{\mbox{vol}_d(\delta \mathbb{B}_r)},
$$
Also,
$$
\E_{U\sim \mathbb{S}_{r}}[C(K_1+U, \ldots K_{n_s}+U) {[U,\ldots,U]}]$$ $$= r\E_{U\sim \mathbb{S}_{r}}[C(K_1+U, \ldots K_{n_s}+U) \frac{[U,\ldots,U]}{r}] $$ $$= r\cdot \frac{\int_{\delta \mathcal{S}_r} C(K_1+U, \ldots K_{n_s}+U)\frac{[U,\ldots,U]}{\|U\|_F} dx}{\mbox{vol}_{d-1}(\delta \mathbb{S}_r)}.
$$
The Lemma follows from combining these equations, and use the fact that
$$
{\mbox{vol}_d(\delta \mathbb{B}_r)} = \mbox{vol}_{d-1}(\delta \mathbb{S}_r)\cdot \frac{r}{d}.
$$
From this lemma we can see that we only need polynomial number of sample to estimate the gradient
\end{proof}

Now we will define different estimate of gradients. 
\begin{equation}\label{eq:finite_m}
\hat{\nabla} = (\hat{\nabla}_1, \ldots, \hat{\nabla}_{n_s})
\end{equation}
where :\\
$$
\widehat{\nabla}_p = \frac{1}{m} \sum_{i=1}^{m}\frac{d.C_i}{r^2}
$$
and :\\
$$
C_i = \sum_{t=0}^{\infty}C(K_{w(t)}+U_i)* U_i * \1{ \omega(t) = p}
$$

\begin{equation}\label{eq:finite_ml}
\Tilde{\nabla} = (\Tilde{\nabla}_1, \ldots, \Tilde{\nabla}_{n_s})
\end{equation}

where :\\
$$
\Tilde{\nabla}_p = \frac{1}{m} \sum_{i=1}^{m}\frac{d.C_i}{r^2}
$$
and :\\
$$
C_i = \sum_{t=0}^{\it -1}C(K_{w(t)}+U_i)* U_i * \1{ \omega(t) = p}
$$
\begin{equation}\label{eq:finite_cost}
\nabla' = (\nabla_1', \ldots, \nabla_{n_s}')
\end{equation}
where:\\
$$
\nabla_p' = \frac{1}{m} \sum_{i=1}^{m}\frac{d.C_i}{r^2}
$$
and
$$
C_i = \sum_{j=0}^{\infty}C^{(l)}(K_{w(j)}+U_i)* U_i * \1{ \omega(j) = p}
$$
(where $C^{(l)}$ is same as defined in lemma \ref{lemma:costfinite})\\
In above definition the $\hat {\nabla}$ is defined as estimated gradient in case of finite trajectories , the $\tilde {\nabla}$ is defined as estimated gradient in case of finite trajectories with finite roll out length. $\nabla_p'$ is the estimated value of $\tilde {\nabla}$.

\begin{restatable}{lem}{deltahatapprox}\label{lemma:deltahatapprox}
for an $ \eps $ there are fixed polynomials $f_r(\frac{1}{\eps})$, $f_{sample}(d,\frac{1}{\eps})$ such that when $r \le \frac{1}{f_r(\frac{1}{\eps})}$ with $m \ge f_{sample}(d,\frac{1}{\eps})$, with high probability(at least $1-(\frac{d}{\eps})^{-d}$ the $\hat{\nabla}$ value is $\eps$ close to $\nabla C(K)$ in Frobenius norm.
\end{restatable}
\begin{proof}
We will break the proof in the two parts as follows:\\
$$
\hat{\nabla} - \nabla C(K) = (\nabla C_r(K) - \nabla C(K))+ (\hat{\nabla}-\nabla C_r(K))
$$
for first term we will choose $f_r(\frac{1}{\eps}) = \min\{1/r_0, 2 g_{diff}/\epsilon\}$ ($r_0$ is chosen later) \\
By lemma (\ref{lemma:delCKperturb}) when r is smaller then $1/f_r(1/\epsilon)=\epsilon/2 g_{diff}$, Every point on sphere have $$\|\nabla C(K+U)-\nabla C(K)\|_F \le \epsilon/4$$
Since $\nabla C_r(K)$ is the expectation of $\nabla C(K+U)$, by triangle inequality $$\|\nabla C_r(K)-\nabla C(K)\|_F \le \epsilon/2$$
Now for second term:\\
From lemma \ref{lemma:stokes} we know that :\\
$$\E[\hat{\nabla}] = \nabla C_r(K)$$
Also each individual sample has
norm bounded by $2dC(K)/r$, so by Vector Bernstein's Inequality
know with $m \ge f_{sample}(d,1/\epsilon) =
\Theta\left(d\left(\frac{dC(K)}{\epsilon r}^2\right)\log
  {d/\epsilon}\right)$ samples, with high probability (at least
$1-(d/\epsilon)^{-d}$) $$\|\hat{\nabla} - \E[\hat{\nabla}]\|_F \le
\epsilon/2$$
Adding the both term proves our claim.\\
$r_0$ is chosen according to the lemma (\ref{lemma:delCKperturb}) in relevant polynomial factors.

\end{proof}

\begin{restatable}{lem}{tildadeltaapprox}\label{lemma:tildadeltaapprox}
for $x \sim \cD$ and $\|x\| \le  L$ , there are polynomials $f_{\it, grad}(d,1/\epsilon)$,
$f_{r,trunc}(1/\epsilon)$,
$f_{sample,trunc}(d,1/\epsilon,\sigma,L^2/\mu)$ such that when $m
\ge f_{sample,trunc}(d,1/\epsilon,L^2/\mu)$, $\ell \ge f_{\it,
  grad}(d,1/\epsilon)$, let $x^i_j, u^i_j (0\le j\le \it)$ be a single
path sampled using our algorithm them
$$ \Tilde{\nabla} - \nabla C(K) \le \eps $$
\end{restatable}
\begin{proof}
First we will divide proof in three parts 
$$
\tilde{\nabla} - \nabla C(K) = (\tilde{\nabla} - \nabla') + (\nabla' - \hat{\nabla}) + (\hat{\nabla}-\nabla C(K))
$$
We have already defined $\nabla'$. \\
For getting the  bound on the term $(\hat{\nabla}-\nabla C(K))$ we will use the lemma \ref{lemma:deltahatapprox} and choose functions as follows: \\
$$h_{r,trunc}(1/\epsilon) = h_r(2/\epsilon)$$
and making sure
$$h_{sample,trunc}(d,1/\epsilon) \ge h_{sample}(d,2/\epsilon)$$
By choosing these we can see that the $(\hat{\nabla}-\nabla C(K))$ is less then $\eps/2$

Now for the term $(\nabla' - \hat{\nabla})$, first we will use the lemma \ref{lemma:costfinite} From this lemma for any $K^{'}$, if we choose :
$$
\it \ge \frac{16d^2\cdot
  C^2(K)(\|Q\|+\|R\|\|K\|^2)}{\epsilon r\mu \sigma_{min}^2(Q)} =:
f_{\it, grad}(d,1/\epsilon)
$$
then it holds that \\
$$
\|C^{(\it)}(K')-C(K')\| \le \frac{r\epsilon}{4d}
$$
multiplying by $\frac{d}{r}$ on both sides, then multiply and divide by term $r$ and $U$ on left side(forbenius norm is same). After that taking average over m and using triangle inequality, we will get\\
$$
\|\frac{1}{m}\sum_{i=1}^{m} \frac{d}{r^2} [\ C^{(\it)}(K+U_i)U_i-C(K+U_i)U_i]\ \|  \le \frac{\epsilon}{4}
$$
$$= \| \nabla' - \hat{\nabla}  \| \le \frac{\epsilon}{4} $$
The remaining first term $\tilde{\nabla} - \nabla')$ , we can see that
$$ \E[\tilde{\nabla}] = \nabla'$$
Now the randomness in $\tilde{\nabla}$ occur due the intial state $x_0$, we are taking expectation over $x_0$ to estimate the $\nabla'$.
Since we have assumed that the  $\|x^i_0\| \le L$, So 
$$(x^i_0)(x^i_0)^\top \preceq
\frac{L^2}{\mu} \E[x_0x_0^\top]$$
We will apply this to the cost and summing over time, we will have
$$
[\sum_{j=0}^{\it-1}(x^i_j)^\top Q_{\omega(j)}x^i_j+(u^i_j)^\top R_{\omega(j)} u^i_j] \le \frac{L^2}{\mu} C(K+U_i).
$$
left side has cost of single trajectory while right side have the bound on it. Now if we take the sum of all trajectories the right hand side will become the $\tilde{\nabla}$. So now we have bound on  $\tilde{\nabla}$ and also $\nabla'$ is expectation of $\tilde{\nabla}$. Here we can use the Vector Bernstein's inequality 
When we have $f_{sample,trunc}(d,1/\epsilon,L^2/\mu)$ is a large enough
polynomial, $\|\tilde{\nabla} - \nabla'\| \le \epsilon/4$ with high
probability.
Summing over all terms:\\
$$
\tilde{\nabla} - \nabla C(K) = (\tilde{\nabla} - \nabla') + (\nabla' - \hat{\nabla}) + (\hat{\nabla}-\nabla C(K))
$$
$$
\le \frac{\eps}{4} + \frac{\eps}{4}+ \frac{\eps}{2} = \eps
$$

\end{proof}

\subsection{Convergence of Gradient descent}
Using the lemmas in previous sections we will now prove the convergence of natural gradient descent \\
First we will show that we can estimate the $X^K$ accurately and then the convergence.
Recall that:
$$
X^K = \left( \sumtinf X_1(t), \ldots, \sumtinf X_{n_s}(t) \right)
$$
$$
\hat{X}^{K} = \textup{diag}\left( \sumtinf X_1(t), \ldots, \sumtinf X_{n_s}(t) \right)
$$
Now define
$$
\tilde{X}^K = (\tilde{X}^{K}_{1}, \ldots, \tilde{X}^{K}_{n_s})
$$
where:
$$
\tilde{X}^{K}_{p} = \frac{1}{m} \sum_{i=1}^{m} \sum_{t=0}^{\it-1} x^i_t (x^i_t)^\top  \1{ \omega(t) = p}
$$


\begin{equation}\label{eq:etagrad}
eta_{grad} = 2\left( \|R\|_{max}+\|B\|^2_{max} \left(1+\frac{2 \xi}{\|B\|_{max}}\frac{\alpha}{\mu}\right) \right) \frac{\alpha}{\Lambda_{min}(Q)}
\end{equation}
where :
$$
\xi = \frac{1}{\Lambda_{min}(Q)} \left( \frac{1+\|B\|^2_{max}}{\mu}\alpha+\|R\|_{max} \right) -1
$$
$\alpha$ is defined as follows:
$$
\mathcal{K}_\alpha \doteq \{K \in \mathcal{K}:C(K)<\alpha \}
$$
for every $\alpha\ge C(K^*)$
\begin{restatable}{lem}{GDKnown}\label{lemma:GDKnown}
Suppose \( C(\hat{K}^0) \) is finite. For any stepsize $\eta \le (eta_{grad})^{-1}$, where $eta_{grad}$ is defined in \ref{eq:etagrad},
the  policy gradient method $\hat{K}_{t+1}= \hat{K}_t-\eta \nabla C(\hat{K}_t) $ converges to the global minimum $\hat{K}^*\in \mathcal{K}$ linearly as follows
\begin{align}
\label{eq:mainCon}
C(\hat{K}_t) - C(\hat{K}^*) &\le \left( 1 - \frac{2\eta \mu^2  \Lambda_{\min}({R})}{\| X^{\hat{K}^*} \|} \right)^t\times \nonumber \\
& \qquad \quad \left(C(\hat{K}_0) - C(\hat{K}^*) \right).
\end{align}
and
\begin{align*}
C(\hat{K}_{t+1}) - C(\hat{K}^*) &\leq \left( 1 -  \frac{2\eta\mu^2\Lambda_{\min}({R})}{\| X^{\hat{K}^*} \|} \right) \times \\
&\qquad \qquad \left( C(\hat{K}_t) - C(\hat{K}^*) \right).
\end{align*}
\end{restatable}

\begin{proof}
This lemma is taken from the \cite{Porto et al.} theorem 1 and lemma 5. \\

\end{proof}

\begin{theorem}\label{GDforMJLS}
Suppose $C(\hat{K}_0)$ is finite and $\mu \ge 0$, Then the update rule
$$
\hat{K}_{t+1} = \hat{K}_t - \eta \nabla C(\hat{K}_t)
$$
with step size
$$
\eta \le \frac{1}{eta_{grad}}
$$
the gradient is estimated using lemma (\ref{lemma:deltahatapprox}), (\ref{lemma:tildadeltaapprox})  with 
$$r = 1/f_{GD, r}(1/\epsilon)$$ and $$m \ge f_{GD, sample}(d,
1/\epsilon, L^2/\mu)$$ samples, both are truncated to
$f_{GD,\it}(d,1/\epsilon)$ iterations, then with high probability
(at least $1-\exp(-d)$) in $T$ iterations where
\[
T>\frac{\|X^{K^*} \|}{\mu \Lambda_{\textrm{min}}(R) } \
\left({eta_{grad}}\right)\log \frac{2(C(\hat{K_0}) -C(\hat{K^*}))}{3 \eps}
\] Where the $eta_{grad}$ is defined in the \ref{eq:etagrad}\\
The natural gradient descent update rule satisfies:\\
$$
C(\hat{K}_T) - C(\hat{K}^*) \le \epsilon
$$

\end{theorem}
\begin{proof}
Proof of this theorem in easy and quite similar to Natural Gradient Descnet where we have to bound one less term and is as follows.

$$
\hat{K_0} = (K_{0,1},\ldots,K_{0,n_s})
$$
$$
\hat{K_t} = (K_{t,1},\ldots,K_{t,n_s})
$$
$$
\hat{K}_{t+1} = \hat{K}_t - \eta \nabla C(\hat{K}_t) 
$$
$$
\hat{K}^{'}_{t+1} = \hat{K}_t - \eta \tilde{\nabla}
$$
$$
\hat{K}^* = (K_{1}^*,\ldots,K_{n_s}^*)
$$
where $\hat{K}^*$ is optimal controller.

Now from lemma \ref{lemma:GDKnown} we know that 
for 
$$
\eta \le \frac{1}{eta_{grad}}
$$
the 
\begin{align*}
C(\hat{K}_{t+1}) - C(\hat{K}^*) &\leq \left( 1 -  \frac{2\eta\mu^2\Lambda_{\min}({R})}{\| X^{\hat{K}^*} \|} \right) \times \\
&\qquad \qquad \left( C(\hat{K}_t) - C(\hat{K}^*) \right)
\end{align*}
So, If we can prove that
$$
C(\hat{K}_{t+1}) - C(\hat{K}^{'}_{t+1}) \le \frac{\epsilon}{2} \frac{\eta \Lambda_{min}(R) \mu^2}{\|X^{\hat{K}^*}\|}
$$
Then, From above two equations we can see that if $C(\hat{K}_t)-C(\hat{K}^* )\ge \eps$, then
\begin{align}\label{onestepgrad}
C(\hat{K}^{'}_{t+1})-C(\hat{K}^*) \le \left( 1-\frac{3 \eta\mu^2\Lambda_{\min}({R})}{2\| X^{\hat{K}^*} \|} \right) \times \left( C(\hat{K}_t) - C(\hat{K}^*) \right)
\end{align}
Now We will use the induction for T steps: \\
We can from the equation \ref{onestepgrad} than the cost value is shrinking as t increasing so, $ C(\hat{K}_t) \le C(\hat{K}_0)  $ \\
If 
$$T \ge \frac{\|X^{\hat{K}^*}\|}{\mu} \,
\left(eta_{grad} \right) 
\, \log \frac{2(C(\hat{K_0}) -C(\hat{K^*}))}{3\eps} $$
Then 
$$
C(\hat{K}_T) - C(\hat{K}^*) \le \eps
$$
Where last steps follows from induction and lemma (\ref{lemma:GDKnown})

So Now we have to prove only below equation and we are done.
\begin{equation}\label{eqn:Costdiffgrad}
C(\hat{K}_{t+1}) - C(\hat{K}^{'}_{t+1}) \le \frac{\epsilon}{2} \frac{\eta \Lambda_{min}(R) \mu^2}{\|X^{\hat{K}^*}\|}
\end{equation}

Here we will use the lemma (\ref{lemma:CKperturb}) for going from $C(K)$ to $K$ \\
so if \eqref{eqn:Costdiffgrad} is true then using lemma (\ref{lemma:CKperturb})
\begin{equation}\label{Kdiffgrad}
    \|\hat{K}^{'}_{t+1} - \hat{K}_{t+1}\|
\le \frac{\epsilon}{2 c_{diff}}\eta \Lambda_{\textrm{min}}(R)
\frac{\mu^2}{\|X^{\hat{K}^*}\|}
\end{equation}
Now definition of $\hat{K}^{'}_{t+1}, \hat{K}_{t+1}$ then the \eqref{Kdiffgrad} will become :\\
$$
\| \tilde{\nabla} - \nabla C(\hat{K}) \| \le \frac{\epsilon}{2 c_{diff}} \Lambda_{\textrm{min}}(R)
\frac{\mu^2}{\|X^{\hat{K}^*}\|}
$$

Using the lemma (\ref{lemma:tildadeltaapprox}) This can be done by choosing :\\
$$
f_{GD,r}(1/\epsilon)
=
f_{r,trunc}(\frac{2c_{diff}\|X^{\hat{K}^*}\|}{\mu^2\Lambda_{min}(R)\epsilon})$$
$$
f_{GD, sample}(d,1/\epsilon,L^2/\mu) = f_{sample,trunc}(d,
\frac{2c_{diff}\|X^{\hat{K}^*}\|}{\mu^2\Lambda_{min}(R)\epsilon},L^2/\mu)
$$
$$f_{GD, \it}(d,1/\epsilon) = f_{\it, grad}(d,
\frac{2c_{diff}\|X^{\hat{K}^*}\|}{\mu^2\Lambda_{min}(R)\epsilon})
$$

and if we choose thses values then the difference between the estimated and exact grad becomes
$$\|\tilde{\nabla} - \nabla C(\hat{K}) \| \le \frac{\epsilon}{2c_{diff}} \Lambda_{\textrm{min}}(R)
\frac{\mu^2}{\|X^{\hat{K}^{*}}\|}
$$

so the equation 3.18 :\\
$$
\|\hat{K}^{'}_{t+1} - \hat{K}_{t+1}\|
\le \frac{\epsilon}{2c_{diff}}\eta \Lambda_{\textrm{min}}(R)
\frac{\mu^2}{\|X^{\hat{K}^*}\|}
$$
If above bound satisfies then our desired bound:
$$
C(\hat{K}_{t+1}) - C(\hat{K}^{'}_{t+1}) \le \frac{\epsilon}{2} \frac{\eta \Lambda_{min}(R) \mu^2}{\|X^{\hat{K}^*}\|}
$$
Will also satisfies and our proof is completed.

\end{proof}

\subsection{Convergence of Natural Gradient Descent}

In this section we will prove the convergence of Gradient descent.

\begin{restatable}{lem}{chiKestimation}\label{lemma:chiKestimation}
for  $x\sim\cD$, $\|x\| \le L$ , $x^1_0, ..., x^m_0$ and $m$ random perturbations $U_i \sim \mathbb{S}_r$ where  r is taken such that  $r \le 1/f_{r,var}(1/\epsilon)$, if we simulate trajectories using these initial points and Algorithm 2 for roll out length $\it \ge f_{\it,
  var}(d,1/\epsilon)$ for m times s.t. $m \ge
f_{varsample, trunc}(d,1/\epsilon, L^2/\mu)$ and estimate the $\widehat{X}^K $
 then almost surely there exists polynomials $f_{r, var}(1/\epsilon)$, $f_{varsample,
  trunc}(d,1/\epsilon,L^2/\mu)$ and $f_{\it, var}(d,1/\epsilon)$, such that with igh probability (at least
$1-(d/\epsilon)^{-d}$) the estimate $\tilde{X}^K$ will satisfies 
$$
\|\tilde{X}^K - X^K \| \le \epsilon
$$
\end{restatable}

\begin{proof}
$X^{K,\it}$ is same as defined as in lemma \ref{lemma:chiKperturbation}
Now let
$$
\overline{X}^K = \frac{1}{m} \sum_{i=1}^{m}  X^{K+U_i}
$$
$$
\overline{X}^{K,\it}= \frac{1}{m} \sum_{i=1}^{m}  X^{K+U_i,\it}
$$
We will divide the proof into three parts as follows
$$
\tilde{X}^K - X^K = (\tilde{X}^K - \overline{X}^{K,\it}) + (\overline{X}^{K,\it} - \overline{X}^K) + (\overline{X}^K - X^K )
$$

For the second term $\overline{X}^{K,\it} - \overline{X}^K$ we will directly use the lemma \ref{lemma:chiKperturbation} and if we choose 
$$
\it \ge f_{\it, var}(d,1/\epsilon) = \frac{8d\cdot C^2(K)}{\epsilon\mu \Lambda_{min}^2(Q)}
$$ 
then the term $\overline{X}^{K,\it} - \overline{X}^K$ is bounded by $\frac{\eps}{4}$

Now for the first term $(\tilde{X}^K - \overline{X}^{K,\it})$ ; we can notice that $\E[\tilde{X}^K ] =
\overline{X}^{K,\it}$ 
Where the expectation is taken over the initial points.\\
Here we will give the similar arguments as we given in the previous lemma for proving the bound for gradient estimated and expectation of estimated gradient.
since: $\|x^i_0\| \le L$ and $(x^i_0)(x^i_0)^\top \preceq
\frac{L^2}{\mu} \E[x_0x_0^\top]$
now for every i in $n_s$ we will have:
$$
\sum_{j=0}^{\it-1}x^i_j(x^i_j)^\top \1{ \omega(j) = i} \preceq \frac{L^2}{\mu}{X^K_i}.
$$
by summing over all m,  we will  have bound on $\tilde{X}^K $ so we can now we can use the $f_{varsample,trunc}$ large enough polynomial such that 
$\|\tilde{X}^K - \overline{X}^{K,\it}\| \le \epsilon/2$ is true with high probability. Here we have used the  Vector Bernstein’s inequality in last step to bound the difference of value and its expectation.\\

For the third term $(\overline{X}^K - X^K) $ we will use the condition proved in the lemma \ref{lemma:delCKperturb} (perturbation results).
When  $$r \le \epsilon\cdot \left(\frac{\Lambda_{\min}(Q) }{C(K)} \right)^2
\frac{\mu}{16 \|B\|_{max} \left(\|A-B K\|_{max}+ 1\right)}$$, 
$$\|X^{K+U_i} - X^K\|\le \epsilon/4$$ Since $\overline{X}^K$ is
the average of $X^{K+U_i}$, by the triangle inequality, 
$\|\overline{X}^K - X^K\| \le \epsilon/4$. 
So all three terms: \\
$$
\tilde{X}^K - X^K = (\tilde{X}^K - \overline{X}^{K,\it}) + (\overline{X}^{K,\it} - \overline{X}^K) + (\overline{X}^K - X^K )
$$
$$
\le  \frac{\eps}{2} + \frac{\eps}{4}+ \frac{\eps}{4} = \eps
$$

\end{proof}

\begin{restatable}{lem}{NGDknown}\label{lemma:NGDknown}
Suppose \( C(\hat{K}^0) \) is finite. For any stepsize $\eta \le \frac{1}{2\left( \| {R} \| + \frac{\| {B} \|^2 C({K}^0)}{\mu} \right)}$,
the  natural policy gradient method $\hat{K}_{t+1}= \hat{K}_t-\eta \nabla C(\hat{K}_t) (\chi^{\hat{K}_t})^{-1}$ converges to the global minimum $\hat{K}^*\in \mathcal{K}$ linearly as follows
\begin{align}
\label{eq:mainCon2}
C(\hat{K}_t) - C(\hat{K}^*) &\le \left( 1 - \frac{2\eta\mu  \Lambda_{\min}({R})}{\| X^{\hat{K}^*} \|} \right)^t\times \nonumber \\
& \qquad \quad \left(C(\hat{K}_0) - C(\hat{K}^*) \right).
\end{align}
and

\begin{align*}
C(\hat{K}_{t+1}) - C(\hat{K}^*) &\leq \left( 1 -  \frac{2\eta\mu\Lambda_{\min}({R})}{\| X^{\hat{K}^*} \|} \right) \times \\
&\qquad \qquad \left( C(\hat{K}_t) - C(\hat{K}^*) \right).
\end{align*}
\end{restatable}
\begin{proof}
This lemma is taken from the \cite{Porto et al.} theorem 3 and lemma 9. \\
For this the author have first proved  some bound for one step then use the induction to get bound for t steps.\\
\end{proof}

\begin{theorem}\label{NGDforMJLS}
Suppose $C(\hat{K}_0)$ is finite and $\mu \ge 0$, Then the update rule
$$
\hat{K}_{t+1} = \hat{K}_t - \eta \nabla C(\hat{K}_t) (\chi^{\hat{K}_t})^{-1}
$$
with step size
$$
\eta \le \frac{1}{2\left( \| {R} \| + \frac{\| {B} \|^2 C({K}_0)}{\mu} \right)}
$$
the gradient and $X^K$ are estimated using lemma (\ref{lemma:deltahatapprox}), (\ref{lemma:tildadeltaapprox}) and (\ref{lemma:chiKestimation})  with 
$$r = 1/f_{NGD, r}(1/\epsilon)$$ and $$m \ge f_{NGD, sample}(d,
1/\epsilon, L^2/\mu)$$ samples, both are truncated to
$f_{NGD,\it}(d,1/\epsilon)$ iterations, then with high probability
(at least $1-\exp(-d)$) in $T$ iterations where
\[
T>\frac{\|X^{K^*} \|}{\mu} \,
\left(\frac{\|R\|}{\Lambda_{\textrm{min}}(R)} + 
\frac{\|B\|^2 C(\hat{K_0})}{\mu \Lambda_{\textrm{min}}(R)} \right) 
\, \log \frac{2(C(\hat{K_0}) -C(\hat{K^*}))}{3 \eps}
\]
The natural gradient descent update rule satisfies:\\
$$
C(\hat{K}_T) - C(\hat{K}^*) \le \epsilon
$$

\end{theorem}
\begin{proof}
$$
\hat{K_0} = (K_{0,1},\ldots,K_{0,n_s})
$$
$$
\hat{K_t} = (K_{t,1},\ldots,K_{t,n_s})
$$
$$
\hat{K}_{t+1} = \hat{K}_t - \eta \nabla C(\hat{K}_t) (\chi^{\hat{K}_t})^{-1}
$$
$$
\hat{K}^{'}_{t+1} = \hat{K}_t - \eta \tilde{\nabla} (\tilde{X}^{\hat{K}_t})^{-1}
$$
$$
\hat{K}^* = (K_{1}^*,\ldots,K_{n_s}^*)
$$
where $\hat{K}^*$ is optimal controller.

Now from lemma \ref{lemma:NGDknown} we know that 
for 
$$
\eta \le \frac{1}{2\left( \| {R} \| + \frac{\| {B} \|^2 C({K}_0)}{\mu} \right)}
$$
the 
\begin{align*}
C(\hat{K}_{t+1}) - C(\hat{K}^*) &\leq \left( 1 -  \frac{2\eta\mu\Lambda_{\min}({R})}{\| X^{\hat{K}^*} \|} \right) \times \\
&\qquad \qquad \left( C(\hat{K}_t) - C(\hat{K}^*) \right)
\end{align*}
So, If we can prove that
$$
C(\hat{K}_{t+1}) - C(\hat{K}^{'}_{t+1}) \le \frac{\epsilon}{2} \frac{\eta \Lambda_{min}(R) \mu}{\|X^{\hat{K}^*}\|}
$$
Then, From above two equations we can see that if $C(\hat{K}_t)-C(\hat{K}^* )\ge \eps$, then
\begin{align}\label{onestep}
C(\hat{K}^{'}_{t+1}) -  C(\hat{K}^*) \le \left( 1 -  \frac{3 \eta\mu\Lambda_{\min}({R})}{2\| X^{\hat{K}^*} \|} \right) \times
 \left( C(\hat{K}_t) - C(\hat{K}^*) \right)
\end{align}
Now We will use the induction for T steps: \\
We can from the equation \eqref{onestep} than the cost value is shrinking as t increasing so, $ C(\hat{K}_t) \le C(\hat{K}_0)  $ \\
If 
$$T \ge \frac{\|X^{\hat{K}^*}\|}{\mu} \,
\left(\frac{\|R\|}{\Lambda_{\textrm{min}}(R)} + 
\frac{\|B\|^2 C(\hat{K_0})}{\mu \Lambda_{\textrm{min}}(R)} \right) 
\, \log \frac{2(C(\hat{K_0}) -C(\hat{K^*}))}{3\eps} $$
Then 
$$
C(\hat{K}_T) - C(\hat{K}^*) \le \eps
$$
Where last steps follows from induction and lemma (\ref{lemma:NGDknown})

So Now we have to prove only below equation and we are done.
\begin{equation}\label{eqn:Costdiff}
C(\hat{K}_{t+1}) - C(\hat{K}^{'}_{t+1}) \le \frac{\epsilon}{2} \frac{\eta \Lambda_{min}(R) \mu}{\|X^{\hat{K}^*}\|}
\end{equation}

Here we will use the lemma (\ref{lemma:CKperturb}) for going from $C(K)$ to $K$ \\
so if \eqref{eqn:Costdiff} is true then using lemma (\ref{lemma:CKperturb})
\begin{equation}\label{Kdiff}
    \|\hat{K}^{'}_{t+1} - \hat{K}_{t+1}\|
\le \frac{\epsilon}{2 c_{diff}}\eta \Lambda_{\textrm{min}}(R)
\frac{\mu}{\|X^{\hat{K}^*}\|}
\end{equation}
Now definition of $\hat{K}^{'}_{t+1}, \hat{K}_{t+1}$ then the \eqref{Kdiff} will become :\\
$$
\| \tilde{\nabla} (\tilde{X}^{\hat{K}})^{-1} - \nabla C(\hat{K}) (X^{\hat{K}})^{-1}\| \le \frac{\epsilon}{2 c_{diff}} \Lambda_{\textrm{min}}(R)
\frac{\mu}{\|X^{\hat{K}^*}\|}
$$
to prove this we will break the right side into two parts
\begin{align}\label{eq:34}
\|\tilde{\nabla}(\tilde{X}^{\hat{K}})^{-1} - \nabla C(\hat{K}) ({X}^{\hat{K}})^{-1}&\| \le  \|\tilde{\nabla}-\nabla\|\|(\tilde{X}^{\hat{K}})^{-1}\| + \|\nabla C(\hat{K})\| \|(\tilde{X}^{\hat{K}})^{-1}-({X}^{\hat{K}})^{-1}\|
\end{align}

For first term for large enough samples from Weyl's theorem: $ \|(\tilde{X}^{\hat{K}})^{-1}\| \le 2/\mu$ 
Now we need to make sure that the
$$\|\tilde{\nabla}-\nabla\|
\le \frac{\epsilon }{8 c_{diff}} \Lambda_{\textrm{min}}(R)
\frac{\mu^2}{\|X^{\hat{K}^*}\|}$$
Using the lemma (\ref{lemma:tildadeltaapprox}) This can be done by choosing :\\
$$
f_{NGD,grad,r}(1/\epsilon)
=
f_{r,trunc}(\frac{8c_{diff}\|X^{\hat{K}^*}\|}{\mu^2\Lambda_{min}(R)\epsilon})$$
$$
f_{NGD, gradsample}(d,1/\epsilon,L^2/\mu) = f_{sample,trunc}(d,
\frac{8c_{diff}\|X^{\hat{K}^*}\|}{\mu^2\Lambda_{min}(R)\epsilon},L^2/\mu)
$$
$$f_{NGD, \it, grad}(d,1/\epsilon) = f_{\it, grad}(d,
\frac{8c_{diff}\|X^{\hat{K}^*}\|}{\mu^2\Lambda_{min}(R)\epsilon})
$$

for Second term:\\
$$\|(\tilde{X}^{\hat{K}})^{-1} - ({X}^{\hat{K}})^{-1}\| \le \frac{\epsilon}{4c_{diff}}
\Lambda_{\textrm{min}}(R) \frac{\mu}{\|X^{\hat{K}^*}\|\|\nabla
  C(\hat{K})\|}$$

Here we will use some equations of standard matrix perturbations.\\
$$\Lambda_{min}(X^{\hat{K}}) \ge \mu
$$ 
$$\|\tilde{X}^{\hat{K}} - {X}^{\hat{K}}\|\le \mu/2$$
$$\|(\tilde{X}^{\hat{K}})^{-1} - ({X}^{\hat{K}})^{-1}\| \le 2\|\tilde{X}^{\hat{K}} - {X}^{\hat{K}}\|/\mu^2
$$
Now for this bound we will use lemma (\ref{lemma:chiKestimation}) and choose:\\
$$f_{NGD, var, r}(1/\epsilon) = f_{var,
  r}(\frac{8c_{diff}\|X^{\hat{K}^*}\|\|\nabla
  C(\hat{K})\|}{\mu^3\Lambda_{min}(R)\epsilon})
$$  
$$
f_{NGD, varsample}(d,1/\epsilon,L^2/\mu) = f_{varsample,trunc}(d, \\ 
\frac{8c_{diff}\|X^{\hat{K}^*}\|\|\nabla C(\hat{K})\|}{\mu^3\Lambda_{\min}(R)\epsilon},L^2/\mu)
$$

$$f{NGD, \it, var}(d,1/\epsilon) = f_{\it, var}(d,\frac{8c_{diff}\|X^{\hat{K}^*}\|\|\nabla C(\hat{K})\|}{\mu^3\Lambda_{min}(R)\epsilon})$$

Finally if we choose:\\
$$f_{NGD,r} = \max\{f_{NGD,grad, r}, f_{NGD,var, r}\}$$ 
$$f_{NGD,sample} = \max\{f_{NGD,gradsample}, f_{NGD,varsample}\}$$ $$f_{NGD,\it} = \max\{f_{NGD,\it, grad},f_{NGD,\it, var}\}$$

then the equation (\ref{eq:29}) becomes
\begin{align*}
& \|\tilde{\nabla}(\tilde{X}^{\hat{K}})^{-1} - \nabla C(\hat{K}) ({X}^{\hat{K}})^{-1}\| \\
& \le \|\tilde{\nabla}-\nabla\|\|(\tilde{X}^{\hat{K}})^{-1}\| + \|\nabla C(\hat{K})\| \|(\tilde{X}^{\hat{K}})^{-1} - ({X}^{\hat{K}})^{-1}\| \\
&\le \frac{\epsilon }{8 c_{diff}} \Lambda_{\textrm{min}}(R)
\frac{\mu^2}{\|X^{\hat{K}^*}\|} \frac{2}{\mu}+ \\
& \frac{\epsilon}{4c_{diff}}
\Lambda_{\textrm{min}}(R) \frac{\mu}{\|X^{\hat{K}^*}\|\|\nabla
  C(\hat{K})\|} |\nabla C(\hat{K})\|\\
&= \frac{\epsilon}{2c_{diff}} \Lambda_{\textrm{min}}(R)
\frac{\mu}{\|X^{\hat{K}^*}\|}
\end{align*}

so the equation \ref{Kdiff} :\\
$$
\|\hat{K}^{'}_{t+1} - \hat{K}_{t+1}\|
\le \frac{\epsilon}{2c_{diff}}\eta \Lambda_{\textrm{min}}(R)
\frac{\mu}{\|X^{\hat{K}^*}\|}
$$
If above bound satisfies then our desired bound:
$$
C(\hat{K}_{t+1}) - C(\hat{K}^{'}_{t+1}) \le \frac{\epsilon}{2} \frac{\eta \Lambda_{min}(R) \mu}{\|X^{\hat{K}^*}\|}
$$
Will also satisfies and our proof is completed.

\end{proof}
\bibliographystyle{unsrt}  


\end{document}